\documentclass[runningheads]{llncs}
\usepackage{graphicx}
\usepackage{amsmath}
\usepackage{amssymb}
\usepackage[usenames,dvipsnames]{xcolor}
\usepackage{mathtools}
\usepackage{adjustbox}
\usepackage{booktabs}
\usepackage{amstext} 
\usepackage{array}   
\usepackage{caption} 
\usepackage[shortlabels]{enumitem} 

\usepackage{graphicx,wrapfig,lipsum}
\usepackage[hidelinks]{hyperref}
\usepackage[subtle,tracking=normal]{savetrees}
\newcommand{\new}[1]{#1}

\newcommand{\err}{\delta}
\usepackage{ntheorem}
\newtheorem*{theorem-non}{Theorem}

\begin{document}
\title{Scalable Stochastic Parametric Verification with Stochastic Variational Smoothed Model Checking}
 
\titlerunning{Scalable Stochastic Parametric Verification with SV-smMC}
%

\author{Luca Bortolussi\inst{1} \and
Francesca Cairoli\inst{1} \and
Ginevra Carbone\inst{1} \and
Paolo Pulcini\inst{1}
}

%
%
\institute{University of Trieste, Italy}

\maketitle

\begin{abstract}
Parametric verification of linear temporal properties for stochastic models requires to compute the satisfaction probability of a certain property as a function of the parameters of the model. Smoothed model checking (smMC)~\cite{bortolussi2016smoothed} infers the satisfaction function over the entire parameter space from a limited set of observations obtained via simulation. As observations are costly and noisy, smMC leverages the power of Bayesian learning based on Gaussian Processes (GP), providing accurate reconstructions with statistically sound quantification of the uncertainty. 
In this paper we propose Stochastic Variational Smoothed Model Checking (SV-smMC), which exploits stochastic variational inference (SVI) to approximate the posterior distribution of the smMC problem. The strength and flexibility of SVI, a stochastic gradient-based optimization making inference easily parallelizable and enabling GPU acceleration,  make SV-smMC applicable both to  Gaussian Processes (GP) and Bayesian Neural Networks (BNN). 
SV-smMC extends the smMC framework by greatly improving scalability to higher dimensionality of parameter spaces and larger training datasets, thus overcoming the well-known limits of GP. 
\new{Additionally, we combine the Bayesian quantification of uncertainty of SV-smMC with the Inductive Conformal Predictions framework to provide probabilistically approximately correct point-specific error estimates, with statistical guarantees over the coverage of the predictive error.}
\end{abstract}

\section{Introduction}\label{sec:intro}

Parametric verification of logical properties aims at providing meaningful insights into the behaviour of a system, checking whether its evolution satisfies or not a certain requirement while varying some parameters of the system's model. The requirement is typically expressed as a temporal logic formula.
Stochastic systems, however, require the use of probabilistic model checking (PMC) techniques as the satisfaction of a property is itself a stochastic quantity, facing significant scalability issues.
To ameliorate such problems, statistical model checking (SMC) uses statistical tools to estimate the satisfaction probability of logical properties from trajectories sampled from the stochastic model. These estimates are enriched with probabilistic bounds of the estimation errors.
If the number of sampled trajectories is sufficiently large, the satisfaction probability, estimated as the average of satisfaction on individual runs, will converge to the true probability. 
However, if the parameters of the stochastic model vary, the dynamics of the system will also vary. Therefore, SMC has to be performed from scratch for each set of parameter values, making SMC computationally unfeasible for stochastic parametric verification.
In this regard, the satisfaction probability of a signal temporal logic (STL) requirement over a parametric stochastic model, in particular population Markov chains, has been proved to be a smooth function of the parameters of the model~\cite{bortolussi2016smoothed}. This result enables the use of machine learning techniques to infer an approximation of this function from a limited pool of observations.  
Observations, computed via SMC for a small number of parameter values, are noisy and may be computationally demanding to obtain. This calls for Bayesian approaches, where predictions provide a probabilistic quantification of the predictive uncertainty. 
In this regard, in~\cite{bortolussi2016smoothed} the authors present smoothed model checking (smMC), a fully Bayesian solution based on Gaussian Processes (GP). Since the observation process is non-Gaussian, outputs are in fact realizations of a Bernoulli distribution, exact GP inference is unfeasible. The authors thus resort to the Expectation Propagation (EP) algorithm to approximate the posterior inference. Unfortunately, the cost of EP is cubic in the number of observations used to train the GP, making smMC applicable only to models with a low dimensional parameter space as they require a limited number of training observations. 
In~\cite{piho2021active}, this scalability problem is tackled by using a sparse variational approach to GP inference. Sparsification reduces the computational complexity, by performing inference on a limited set of observations, called inducing points. On the other hand, variational inference is used to perform approximate inference of a GP classification (GPC) problem. The variational approach for GPC used in~\cite{piho2021active} builds on~\cite{titsias2009variational}. Here the objective function does not depend explicitly on the inducing variables, forcing them to be fixed a priori, leaving no room for an optimal selection of such points. Moreover, the loss does not allow the dataset to be divided in mini-batches making stochastic gradient descent not applicable for optimization.
Scalability is thus improved, but sparsification strongly reduces the reconstruction accuracy. Moreover, in~\cite{piho2021active} the smMC problem is framed as a GPC problem, meaning that observations come from the satisfaction of a single trajectory. If for a certain parameter we simulate $M$ trajectories, this would result in $M$ different observations. On the contrary, in~\cite{bortolussi2016smoothed} the observation process was modeled by a binomial so that the satisfaction of the $M$ simulations is condensed into a single observation. This has a strong effect on the dimension of the training set which is of paramount importance for sake of scalability.
Finding an effective solution that makes smMC scale to large datasets remains an open issue. 

\paragraph{Main contributions.} \new{ The first contribution is to tackle the scalability issues of smMC by leveraging SVI, instead of EP, to solve the smMC Bayesian inference problem. More precisely,} we propose a novel approach for scalable stochastic parametric verification, called Stochastic Variational Smoothed Model Checking (SV-smMC), that leverages Stochastic Variational Inference (SVI) to make the smMC Bayesian inference scale to large datasets. 
The variational rationale is to transform the inference problem into an optimization one. The approach is stochastic in the sense that stochastic gradient descent (SGD) is used in the gradient-based optimization of a suitable variational objective function.
SVI is extremely flexible and it is thus applied both to Gaussian Processes (GP) and to Bayesian Neural Networks (BNN). The main advantage of SVI, compared to the VI used for example in~\cite{piho2021active}, is the use of mini-batches that makes inference easily parallelizable, enabling also GPU acceleration. \new{Moreover, in SVI the inducing variables are optimally selected during inference so that sparsification causes a less pronounced drop in the reconstruction accuracy w.r.t.~\cite{piho2021active}}. As a result, SV-smMC can face extremely large datasets.

\new{The second important contribution is that of enriching smMC, either the original EP-based version or the novel SVI-based version, with \textit{probabilistically approximately correct 
 statistical guarantees over the generalization error} holding point-wise at any prediction point. We obtain this result by combining Bayesian estimates of uncertainty with Inductive Conformal Predictions~\cite{vovk2005algorithmic}. } 


\paragraph{Overview of the paper.} This paper is structured as follows. We start by presenting the background theory in Sect.~\ref{sec:background}, comprised of the formal definition of the smMC Bayesian inference problem (Sect.~\ref{subsec:smmc}). In Sect.~\ref{sec:sv-smc} the theoretical details of SV-smMC are presented both for the GP version (Sect.~\ref{subsec:gp-smc}) and for the BNN version (Sect.~\ref{subsec:bnn-smc}). \new{Sect.~\ref{sec:stat_guar} presents Inductive Conformal Predictions and the details of the technique used to have local statistical guarantees over the predictive error.} Finally, in Sect.~\ref{sec:experiments} we compare the performances of SV-smMC against those of smMC on three stochastic models with increasing parametric complexity. Two of these case studies are taken from~\cite{bortolussi2016smoothed} to make a fair comparison of the performances. Moreover, in order to have a better quantification of the scalability of the proposed solution, we test SV-smMC on a pool of randomly generated stochastic processes over parameter spaces of increasing dimension for multiple randomly generated temporal properties.
\section{Background}\label{sec:background}

\subsection{Population Continuous Time Markov Chain}

A population of interacting agents can be modeled as a stochastic system evolving continuously in time over a finite or countable state space $\mathcal{X}$. Assuming the system is Markovian, meaning the memory-less property holds, we can rely on population Continuous Time Markov Chains (CTMC) $\mathcal{M}$. A population is specified by $n$ different species $\{S_1,\dots, S_n\}$ subject to a dynamics described by $r$ different rules (reactions) $\{R_1,\dots, R_r\}$. The respective CTMC is described by:
\begin{itemize}
    \item a state vector, $X(t) = (X_1(t),\dots , X_n(t))$, taking values in $\mathcal{X}\subseteq \mathbb{N}^n$ and counting the number of agents in each species at time $t$; 
    \item a finite set of reactions $R = (R_1, \dots, R_r)$ describing how the state of the population changes in time. A general reaction $R_i$ is identified by the tuple $(\tau_i,\nu_i)$, where:
    \begin{itemize}
        \item[-] $\rho_i:\mathcal{X}\times\Theta_i\to\mathbb{R}_{\ge 0}$ is the \emph{rate function} of reaction $R_i$ that associates with each reaction the rate of an exponential distribution, as a function of the global state of the model and
of parameters $\theta_i$, and
        \item[-] $\nu_i$ is the \emph{update vector}, giving the net change of agents due to the reaction, so that the firing of reaction $R_i$ results in a transition of the system from state $X(t)$ to state $X(t)+\nu_i$.
    \end{itemize}
Reaction rules are easily
visualised in the chemical reaction style, as
$$
R_i: \underset{j\in\{1,\dots, n\}}{\sum}\alpha_{ij}S_j \overset{ \rho_i(X,\theta_i)}{\longrightarrow} \underset{j\in\{1,\dots, n\}}{\sum}\beta_{ij}S_j.
$$
\end{itemize}
The stoichiometric coefficients $\alpha_i = [\alpha_{i1},\dots, \alpha_{in}], \beta_i = [\beta_{i1},\dots, \beta_{in}]$ can be arranged so that they form the update vector $\nu_i = \beta_i-\alpha_i$ for reaction $R_i$.

The parameters $\theta = (\theta_1,\dots,\theta_r)$ have a crucial effect on the dynamics of the system: changes in $\theta$ can lead to qualitatively different dynamics. We stress such crucial dependency by using the notation $\mathcal{M}_\theta$. The trajectories of such a CTMC can be seen as samples of random variables $X(t)$ indexed by time $t$ over a state space $\mathcal{X}$.
A parametric CTMC (pCTMC) is a family $\mathcal{M}_\theta$ of CTMCs where the parameters $\theta$ vary in a domain $\Theta$.

\subsection{Signal Temporal Logic}

Properties of CTMC trajectories can be expressed via Signal Temporal Logic (STL)~\cite{maler2004monitoring} formulas. STL
allows the specification of properties of dense-time, real-valued signals, and the automatic generation of monitors for
testing properties on individual trajectories. The rationale of STL is to transform real-valued signals into Boolean ones, using formulae built on the following \emph{STL syntax}:
\begin{equation}\label{eq:stl_syntax}
    \varphi := true\ |\ \mu\ |\ \neg\varphi\ |\  \varphi\land\varphi\ |\ \varphi\ \mathcal{U}_I\varphi,
\end{equation}
where $\mathcal{U}_I$ is the until operator, $I\subseteq \mathbb{T}$ is a temporal interval, either bounded, $I = [a,b]$, or unbounded, $I = [a,+\infty)$, for any $0\le a < b$. Atomic propositions $\mu$ are (non-linear) inequalities on population
variables. 
From this essential syntax it is easy to define other operators: $false :=\neg true$, $\varphi \lor \psi := \neg(\neg\varphi\land\neg\psi)$, eventually $F_I:= true\ \mathcal{U}_I\varphi$ and globally $G_I:= \neg F_I\neg\varphi$. Monitoring the satisfaction of a formula is done recursively on the parsing tree structure of the STL formula. See~\cite{maler2004monitoring} for the details on STL Boolean semantics and on Boolean STL monitors.

\subsection{Smoothed Model Checking}\label{subsec:smmc}

\paragraph{Probabilistic Model Checking (PMC).}  Verification of temporal properties is of paramount importance, especially for safety-critical processes. When the system evolves stochastically probabilistic model checking~\cite{baier2008principles} comes into play.
For linear time properties, like those of STL, the goal is to compute 
the probability $Pr(\varphi|\mathcal{M})$ that a stochastic system $\mathcal{M}$ satisfies a given STL formula $\varphi$. 
Exact computation of $Pr(\varphi|\mathcal{M})$ suffers from
very limited scalability and furthermore requires the full knowledge of the stochastic model.

\paragraph{Statistical Model Checking (SMC).}
Statistical model checking~\cite{younes2002probabilistic} fights the two aforementioned issues: instead of analyzing the model,
it evaluates the satisfaction of the given formula on a number of observed runs
of the system, and derives a statistical estimate of $Pr(\varphi|\mathcal{M})$, valid only with some
confidence. Given a CTMC $\mathcal{M}_\theta$ with fixed
parameters $\theta$, time-bounded CTMC trajectories are sampled by standard simulation algorithms, such as SSA~\cite{gillespie1977exact}, and monitoring algorithms for STL~\cite{maler2004monitoring} are
used to assess if the formula $\varphi$ is satisfied for each sampled trajectory.
This process produces samples from a Bernoulli random variable equal to $1$ if and only if $\varphi$ is true. SMC~\cite{younes2006statistical,zuliani2010bayesian} then uses standard statistical tools, either frequentist~\cite{younes2006statistical} or Bayesian~\cite{zuliani2010bayesian}, to estimate from these samples the satisfaction probability $Pr(\varphi|\mathcal{M}_\theta)$ or to
test if $Pr(\varphi|\mathcal{M}_\theta) > q$ with a prescribed confidence level.

\paragraph{Satisfaction Function for pCTMCs.}
Building on~\cite{bortolussi2016smoothed}, our interest is in parametric verification~\cite{jansen2014accelerating}, i.e. to quantify how the satisfaction of
STL formulae depend on
the unknown parameters of the pCTMC. We define the \emph{satisfaction function} $f_\varphi : \Theta\to [0,1]$ associated to $\varphi$ as
\begin{equation}\label{eq:sat_fnc}
    f_\varphi (\theta) = Pr(\varphi= true | \mathcal{M}_\theta).
\end{equation}
An accurate estimation of the satisfaction function $f_\varphi$ over the entire parameter space $\Theta$ by means of SMC would require a prohibitively large number of evaluations. In~\cite{bortolussi2016smoothed} -- Theorem 1 -- it has been shown that $f_\varphi (\theta)$ is a smooth function of the model parameters and thus machine learning techniques can be used to infer this function from a limited set of observations and this is exactly the goal of smoothed model checking (smMC).

\paragraph{Smoothed model checking.}
Given a pCTMC $\mathcal{M}_\theta$ and an STL formula $\varphi$, the goal of smMC is to find a statistical estimate of the satisfaction function of~\eqref{eq:sat_fnc} from a set of noisy observations of $f_\varphi$ obtained at few parameter values $\theta_1,\theta_2,\dots$. The task is to construct a statistical model that, for any value $\theta^*\in\Theta$, computes efficiently an estimate of $f_\varphi(\theta^*)$ together with a credible interval associated with the prediction. 
More precisely, given an input point $\theta$, our observations are obtained by evaluating a property $\varphi$ on few trajectories sampled from the stochastic model $\mathcal{M}_\theta$ via SSA. 
Thus, given a set of $N_t$ parameter values, $\Theta_t = \{\theta_1, \dots,\theta_{N_t}\}$, we simulate, for each parameter $\theta_i$, $M_t$ trajectories, obtaining $M_t$ Boolean values $\ell_i^j\in \{0,1\}$ for $j=1,\dots,M_t$. We condense these Boolean values in a vector $L_i = [\ell_i^1, \dots ,\ell_i^{M_t}]$ .
The noisy observations form the \emph{training set}
\begin{equation}\label{eq:train_dataset}
    D_t = \left\{
    (\theta_i, L_i)\mid i = 1,\dots, N_t
    \right\}. 
\end{equation}

In practice, smMC can be framed as a \emph{Bayesian inference problem} that aims at inferring an accurate probabilistic estimate of the unknown satisfaction function $f_\varphi:\Theta\to [0,1]$. In the following, let $f:\Theta \to [0,1]$.
The main ingredients of a Bayesian approach are the following:
\begin{enumerate}

    \item Choose a \emph{prior} distribution, $p(f)$, over a suitable function space, encapsulating the beliefs about function $f$ prior to any observations being taken.
    \item Determine the functional form of the observation process by defining a suitable \emph{likelihood} function that effectively models how the  observations depend on the uncertain parameter $\theta$. Our observation process can be modeled by a binomial over $M_t$ trials with parameter $f_\varphi (\theta)$. Given the nature of our training set, defined in~\eqref{eq:train_dataset}, we define the probabilistic likelihood as 
    $$p(D_t| f) = \prod_{i=1}^{N_t} Binomial(L_i\mid M_t, f(\theta_i)).$$
    
    \item Leverage Bayes' theorem to define the \emph{posterior} distribution over functions given the observations
    $$
    p(f| D_t) = \frac{p(D_t| f)p(f)}{p(D_t)}.
    $$
    Computing $p(D_t) = \int p(D_t| f) p(f) df$ is almost always computationally intractable as we have non-conjugate prior-likelihood distributions. Therefore, we need algorithms to accurately approximate such posterior distribution.
    
    \item Evaluate such posterior at points $\theta_*$, resulting in a predictive distribution $p(f_*|\theta_*, D_t)$, whose statistics are used to obtain the desired estimate of the satisfaction probability together with the respective credible interval.

\end{enumerate}

\noindent Two main ingredients are essential to define the smMC solution strategy: 
\begin{itemize}
    \item[(i)] the probabilistic model chosen to describe the distribution over functions $f$,
    \item[(ii)] the approximate inference strategy.
\end{itemize}

\paragraph{Previous works and limitations.}
The smMC technique presented in~\cite{bortolussi2016smoothed} uses Gaussian Processes (GP) as probabilistic model -- ingredient $(i)$ -- and the Expectation Propagation (EP) algorithm to approximate the posterior distribution -- ingredient $(ii)$. This solution scales as $\mathcal{O}(N_t^3)$, it is thus unfeasible for large datasets. 
Notice that, if the observation process is modeled by a Bernoulli, as in~\cite{piho2021active}, meaning if Boolean values $\ell_i^j$ are considered as individual observations instead of condensing them in a vector $L_i$, inference scales as $\mathcal{O}\left((N_t M_t)^3\right)$. In order to mitigate such scalability issues, in~\cite{piho2021active} the authors propose an alternative for ingredient $(ii)$ using variational inference together with sparsification techniques to make inference feasible for slightly larger datasets. Sparsification reduces the computational complexity to $\mathcal{O}\left((m M_t)^3\right)$, where $m$ is the number of sparse observations considered.  Nonetheless, sparsification strongly
reduces the reconstruction accuracy, \new{especially if the inducing variable are not optimally selected as in~\cite{piho2021active}}. Moreover, the variational optimization problem of the proposed approach cannot be framed in terms of stochastic gradient descent optimization. These issues strongly limit the scalability capabilities of the proposed solution.

\paragraph{Our contribution.} The main contribution of this paper, explained in the next section, is to introduce stochastic variational inference (SVI) as an efficient and scalable alternative for ingredient $(ii)$. The rationale is to frame Bayesian inference as an optimization problem, typical workaround of variational approaches, and define an objective function that enables optimization in terms of stochastic gradient descent (SGD) so that smMC becomes efficient and scalable even on extremely large datasets. SVI is then applied on two different probabilistic models, i.e. on two alternatives for ingredient $(i)$: Gaussian Processes (GP) and Bayesian Neural Networks (BNN).
\new{The second main contribution, presented in Sect.~\ref{sec:stat_guar}, is the calibration of the Bayesian estimates of uncertain in order to obtain point-specific bounds with guaranteed coverage of the predictive error.}

\section{Stochastic Variational Smoothed Model Checking}\label{sec:sv-smc}

The goal of Stochastic Variational Smoothed Model Checking (SV-smMC) is to make smMC scalable to high-dimensional models.
SV-smMC proposes stochastic variational inference as ingredient $(ii)$ to efficiently compute the approximate posterior distribution $p(f|D_t)$ so that smMC inference scales well to large datasets $D_t$. 

The core idea behind variational approaches is to translate the posterior inference into an optimization problem, where a parametric distribution is proposed as a candidate approximator of the unknown posterior distribution. The optimization problem aims at minimizing the difference, measured by the Kullback-Leibler (KL) divergence, between these two distributions. However, as the posterior is unknown, various model-specific strategies can be developed to derive from the KL formula a lower bound of the marginal log-likelihood of our data. This lower bound, known as evidence lower bound (ELBO), is used as new objective function as it does not depend explicitly on the unknown posterior. This bound is then optimized with respect to the parameters of the proposed variational distribution. The latter is typically chosen to have nice statistical properties, making predictions feasible and efficient. Owing to the work of \cite{hoffman2013stochastic}, in order to scale VI over very large datasets, the (black-box VI) optimization task can be phrased as a stochastic optimization problem~\cite{zinkevich2010parallelized} by estimating the gradient of the ELBO with Monte Carlo methods. Moreover, the dataset can be divided into mini-batches. As for ingredient $(i)$, SV-smMC can use two alternative probabilistic models to define distributions over function $f$. The first one is based on Gaussian Processes (GP), whereas the second one is based on Bayesian Neural Networks (BNN). SVI is applied to both probabilistic models with the proper model-specific adjustments to the variational formulation. Below, we provide an intuitive presentation of the approximate inference processes, whereas more formal details are provided in Appendix~\ref{app:svi_details}.

\subsection{Gaussian Processes over non-Gaussian likelihoods}\label{subsec:gp-smc}

\textit{Gaussian Processes (GP)} are a well-known formalism to define distributions over real-valued functions of the form $g: \Theta\rightarrow\mathbb{R}$. A GP distribution is uniquely identified by its mean $\mu(\theta)=\mathbb{E}[g(\theta)]$ and its covariance function $k_\gamma(\theta,\theta')$ and characterized by the fact that the distribution of $g$ over any finite set of points $\hat{\theta}$ is {Gaussian} with mean $\mu(\hat{\theta})$ and variance $k_\gamma(\hat{\theta},\hat{\theta})$.  
In the following, we let $g_t$, $\mu_t$ and $K_{N_tN_t}$ denote respectively the latent, the mean and the covariance functions evaluated on the training inputs $\Theta_t$.

The GP prior over latent functions $g$ evaluated at training inputs $\Theta_t$ -- step 1 -- is defined as $p(g|\Theta_t) = \mathcal{N}(g|\mu_t, K_{N_tN_t})$. The posterior over latent variables $p(g_t|D_t)$ -- step 3 -- is not available in closed form since it is the convolution of a Gaussian and a binomial distribution. 
Hence, we have to rely on SVI for  posterior approximation (details later). Once we obtain a tractable posterior approximation, in order to make predictions over a test input $\theta_*$, with latent variable $g_*$, we have to compute an empirical approximation of the predictive distribution
\begin{equation}\label{eq:pred_gp}
    p(f_*|\theta_*,D_t) = \int \Phi (g_*)p(g_*|\theta_*, D_t)dg_*,
\end{equation}
in which the outputs of the latent function $g:\Theta\to \mathbb{R}$ are mapped into the $[0, 1]$ interval by means of a so-called link function $\Phi$, typically the \new{inverse} logit or the \new{inverse} probit function~\cite{bishop2006pattern}, so that $f:\Theta\to [0,1]$ is obtained as $f = g \circ \Phi$.

\paragraph{Stochastic Variational Inference.}
Here we outline an intuitive explanation of the SVI steps to approximate the GP posterior when the likelihood is non-Gaussian. For a more detailed mathematical description see Appendix~\ref{app:svi_details}. 
The main issue with GP inference is the inversion of the $N_t\times N_t$ covariance matrix $K_{N_tN_t}$. This is the reason why variational approaches to GP start with sparsification, i.e. with the selection of $m \ll N_t$ inducing points that live in the same space of $\Theta_t$ and, from them, define a set of inducing variables $u_t$. The covariance matrix over inducing points, $K_{mm}$, is less expensive to invert and thus it acts as a low-rank approximation of $K_{N_tN_t}$. We introduce a Gaussian variational distribution $q(u_t)$ over inducing variables whose goal is to be as similar as possible to the posterior $p(u_t|D_t)$. A classical VI result is to transform the expression of the KL divergence between the variational distribution $q(u_t)$ and the posterior $p(u_t|D_t)$ into a lower bound over the marginal log-likelihood $\log p(D_t)$. As our
likelihood -- step 2 -- factors as $p(D_t | g_t ) = \prod_{i=1}^{N_t} p(L_i | g_i )$ and because of the Jensen inequality we obtain the following ELBO (see Appendix~\ref{app:svi_gp} for the mathematical details):

\begin{equation}\label{eq:gp_svi_bound}
   \log p(D_t)\ge \sum_{i=1}^{N_t}\mathbb{E}_{q(g_i)}[\log p(L_i|g_i)]-KL[q(u_t)||p(u_t)]:= \mathcal{L}_{GP}(\nu,  \gamma),
\end{equation}
where $L_i$ denotes the set of observed Boolean tuples corresponding to points in $\theta_i$ in $D_t$, $p(u_t)$ denotes the prior distribution over inducing variables and $\nu$ denotes the hyper-parameters introduced to describe the sparsification and the variational distribution. The distribution $q(g_t)$ is Gaussian with an exact analytic derivation from $q(u_t)$ that requires $\mathcal{O}(m^2)$ computations.
The SVI algorithm then consists of maximizing $\mathcal{L}_{GP}$ with respect to its parameters using gradient-based stochastic optimization. \new{We stress that, at this step, the selection of inducing variable is optimized, resulting in a more effective sparsification.} Computing the KL divergence in~\eqref{eq:gp_svi_bound} requires only $\mathcal{O}(m^3)$ computations. Most of the work will thus be in computing the
expected likelihood terms. 
Given the ease of parallelizing
the simple sum over $N_t$, we can optimize $\mathcal{L}_{GP}$ in a stochastic fashion by selecting mini-batches of the data at random.

\paragraph{Predictive distribution.} The predictive posterior $p(g_*|\theta_*, D_t)$ is now approximated by a variational distribution $q(g_*)$, which is Gaussian and whose mean and variance can be analytically computed with cost $\mathcal{O}(m^2)$. From the mean and the variance of $q(g_*)$, we obtain the respective credible interval and we can use the link function $\Phi$ to map it to a subset of the interval $[0,1]$, in order to obtain the mean and the credible interval of the posterior predictive distribution $p(f_*|\theta_*, D_t)$ of equation~\eqref{eq:pred_gp}.

\subsection{Bayesian Neural Networks}\label{subsec:bnn-smc}

The core idea of Bayesian neural networks (BNNs)~\cite{lampinen2001bayesian,titterington2004bayesian,goan2020bayesian,bishop2006pattern}, is to place a
probability distribution over the weights $\mathbf{w}$ of a neural network $f_{\mathbf{w}}:\Theta\to[0,1]$, transforming the latter into a probabilistic model.

The Bayesian learning process starts by defining a prior distribution $p(\mathbf{w})$ over $\mathbf{w}$ -- step 1 -- that expresses our initial belief about the values of the weights. 
As we observe data $D_t$, we update this prior to a posterior distribution $p(\mathbf{w}| D_t)$ -- step 3 -- using Bayes' rule.
Because of the non-linearity introduced by the
neural network function $f_{\mathbf{w}}(\theta)$ and since the likelihood $p(D_t|\mathbf{w})$ -- step 2 -- is binomial, the posterior $p(\mathbf{w}|D_t)$ is non-Gaussian and it cannot be computed analytically. 
In order to predict the satisfaction function over an unobserved input $\theta_*$, we marginalize
the predictions with respect to the posterior distribution
of the parameters, obtaining 
\begin{equation}\label{eq:bnn_predictive}
    p(f_*|\theta_*, D_t) = \int f_\mathbf{w}(\theta_*) p(\mathbf{w}|D_t)d\mathbf{w}.
\end{equation}
The latter is called \emph{posterior predictive} distribution and it can be used to retrieve information about the uncertainty of a specific prediction $f_*$.
Unfortunately, the integration is analytically intractable due to the non-linearity of the neural network function~\cite{bishop2006pattern,mackay1992practical} so we empirically estimate such quantity.

\paragraph{Stochastic Variational Inference.}
 The rationale of SVI for BNN is to choose a parametric variational distribution $q_\psi(\mathbf{w})$ that approximates the unknown posterior distribution $p(\mathbf{w}|D_t)$ by minimizing the KL divergence $KL[q_\psi(\mathbf{w})||p(\mathbf{w}|D_t)]$ between these two distributions. Since the posterior distribution is not known, the classic variational approach is to transform the minimization of the KL divergence into the maximization of the Evidence Lower Bound (ELBO)~\cite{jordan1999introduction}, defined as
\begin{equation}\label{eq:elbo}
    \mathcal{L}_{BNN}(\psi) := \mathbb{E}_{q_\psi(\mathbf{w})}\left[\log p(D_t|\mathbf{w})\right]-KL\left[q_\psi(\mathbf{w})||p(\mathbf{w})\right]\le \log p(D_t),
\end{equation}
see Appendix~\ref{app:svi_bnn} for the mathematical details.
The first term is the expected log-likelihood of our data with respect to values of $f_\mathbf{w}$ sampled from $q_\psi(\mathbf{w}|D_t)$, whereas the second term is the KL divergence between the proposal distribution and the prior. The distribution $q_\psi$ should be a distribution easy to sample from and such that the KL divergence is easy to compute. A common choice for $q_\psi$ is the Gaussian distribution (where $\psi$ denotes its mean and variance). KL divergence among two Gaussian distributions has an exact analytical form, hence the ELBO of~\eqref{eq:elbo} can be computed and it can be used as the objective function of a maximization problem over $\psi$.

\paragraph{Predictive distribution.} The predictive distribution~\eqref{eq:bnn_predictive} is a non-linear combination of Gaussian distributions, and thus it is not Gaussian. However, samples can be easily extracted from $q_\psi(\mathbf{w})$, which allows us to obtain an empirical approximation of the predictive distribution. Let $\it [w_1,\dots, w_C]$ denote a vector of $C$ realizations of the random variable $\mathbf{w}\sim q_\psi(\mathbf{w})$. Each realization $w_i$ induces a deterministic function $f_{w_i}$ that can be evaluated at $\theta_*$, the unobserved input, providing an empirical approximation of $p(f_*|\theta_*, D_t)$.

\new{
\section{Statistical guarantees}\label{sec:stat_guar}

In order to provide provable probabilistic bounds on smMC, we rely on Inductive Conformal Predictions (ICP)~\cite{vovk2005algorithmic,Zeni2020ConformalPA}, a framework that can be applied on top of any deterministic regressor to enrich its predictions with statistically valid quantification of the predictive uncertainty. 
Let $\err:[0,1]^2\to [0,\infty)$ be the \emph{error function} that measures the absolute error made by a predictor $f$ at a point $(\theta_i,f_\varphi(\theta_i))$. Mathematically, $\err\left(f_\varphi(\theta_i),f(\theta_i)\right) = |f_\varphi(\theta_i)-f(\theta_i)|$, where $f_\varphi$ is the satisfaction function. 
The error is bounded and ranges in the interval $[0,1]$.

ICP requires a calibration set $D_c$, meaning an additional batch of $N_c$ observations, coming from the same data generating distribution of $D_t$, but not used during the inference phase. The error function $\delta$ can be used to compute the nonconformity scores over the calibration set: 
$$\tau_c = \big\{\delta\big(f_\varphi(\theta_i),f(\theta_i)\big)\ | \ (\theta_i,f_\varphi(\theta_i))\in D_c\big\}.$$ 
Given a significance level $\varepsilon$, one can extract
$\tau_\varepsilon$, the $(1-\varepsilon$)-quantile\footnote{As the calibration set is finite, $\tau_{\varepsilon}$ is properly defined as the $\big\lceil\frac{(N_c+1)(1-\varepsilon)}{N_c}\big\rceil$-quantile.} of the scores $\tau_c$ over $D_c$.
As ICP is defined for deterministic predictors, we consider predictions made by the pointwise expectation over the posterior $q(f)$ and obtain the following coverage guarantees: 
\begin{equation}\label{eq:icp}
    Pr_{(\theta,L)\sim p(\theta,f_\varphi(\theta))}\Big(
 \delta\big(f_\varphi(\theta),\ \mathbb{E}_{f\sim q(f)}[f(\theta)]\big)\le\tau_\varepsilon
    \Big) \ge 1-\varepsilon,
\end{equation}
where $p(\theta,f_\varphi(\theta))$ is the data generating distribution.
The main limitation of~\eqref{eq:icp} is that the bound $\tau_\varepsilon$ is constant over the entire parameter space $\Theta$, making ICP non-informative to check how the uncertainty distributes over $\Theta$. 

Normalized Inductive Conformal Predictions (NICP)~\cite{Papadopoulos2011ReliablePI,Papadopoulos2014RegressionCP} tackle this limitation. In order to get individual bounds for each point $\theta_i$, one needs normalized nonconformity scores 
\begin{equation}\label{eq:norm_scores}
\tilde{\tau}_c = \Bigg\{\frac{\delta\big(f_\varphi(\theta_i),f(\theta_i)\big)}{u(\theta_i)}\ \Big| \ (\theta_i,f_\varphi(\theta_i))\in D_c\Bigg\},
\end{equation}
where $u(\theta_i)$ should estimate the difficulty of predicting $f_\varphi(\theta_i)$. 
The rationale is that if two points have the same nonconformity scores using $\delta$, the one expected to be more accurate, should be stranger (more nonconforming) than the other one. Hence, we aim at error bounds that are tighter for parameters $\theta$ that are deemed easy to predict and vice-versa. Even for locally-weighted
residuals, as in~\eqref{eq:norm_scores}, the validity of the conformal methods carries over.
As before we compute $\tilde{\tau}_\varepsilon$ as the $(1-\varepsilon)$-quantile of the scores $\tilde{\tau}_c$ and the coverage guarantees over the error become:
\begin{equation}\label{eq:nicp}
    Pr_{(\theta,f_\varphi(\theta))\sim p(\theta,f_\varphi(\theta))}\Big(
 \delta\big(f_\varphi(\theta),\ \mathbb{E}_{f\sim q(f)}[f(\theta)]\big)\le\tilde{\tau}_\varepsilon\cdot u(\theta)
    \Big) \ge 1-\varepsilon.
\end{equation}
The Bayesian quantification of uncertainty, despite being based on statistically sound operations, offers no guarantees per se as it strongly depends on the chosen prior. Additionally, in smMC, either EP- or SVI-based, we further add the error of approximate inference on top of this. However, we can exploit the quantification of uncertainty these methods provide and use it as the normalizing function of a NICP approach, that in turn will provide us with point-specific statistical guarantees over the error coverage. The result for smMC is summarized in the following theorem. 
\begin{theorem}\label{thm:nicp}
Consider a pCTMC $\mathcal{M}_\theta$, an STL formula $\varphi$, a dataset $D_t$ used to infer the smMC posterior distribution $q(f)$ over functions $f$ that approximates the satisfaction function $f_\varphi$, and a calibration set $D_c$. Let the normalizing function $u(\theta)$ be the standard deviation $\sigma_{q(f)}(\theta)$ of the posterior $q(f)$. Then, for a significance level $\varepsilon\in (0,1)$

\begin{equation}\label{eq:thm_main}
    Pr_{(\theta,L)\sim p(\theta,f_\varphi(\theta))}\Big(
 \delta\big(f_\varphi(\theta),\ \mathbb{E}_{f\sim q(f)}[f(\theta)]\big)\le\tilde{\tau}_\varepsilon\cdot \sigma_{q(f)}(\theta)
    \Big) \ge 1-\varepsilon,
\end{equation}
where $\tilde{\tau}_\varepsilon$ is derived according to~\eqref{eq:norm_scores} for $u(\cdot) = \sigma_{q(f)}(\cdot)$.
\end{theorem}

The proof is provided in Appendix~\ref{app:icp}. We remark that the theorem is valid even if a quantile of the posterior $q(f)$ is chosen as a normalizing term.

Theorem~\ref{thm:nicp} provides guarantees over the error made w.r.t. to the ground truth target, the satisfaction function $f_\varphi$. Unfortunately, this quantity cannot be observed exactly. To practically compute a bound, we resort to SMC and estimate it as
$\bar{L}_i: = \tfrac{1}{M_t}\sum_{j_=1}^{M_t}\ell_i^j$, evaluating the nonconformity scores $\tau_c$ as a distance from $\bar{L}_i$ rather than $f_\varphi(\theta_i)$, where $M_t$ has to equal the amount of Bernoulli trials per step for the theorem to remain valid. This introduces an additional error that can be taken into account by combining~\eqref{eq:thm_main} with standard Chernoff bounds for  SMC~\cite{Legay2010StatisticalMC}. Details are reported in Appendix~\ref{app:icp}.
}

\section{Experiments}\label{sec:experiments}

\subsection{Case studies}
\label{sec:case_studies}

We briefly introduce the case studies used to investigate the scalability and the accuracy of SV-smMC. In order to make a fair comparison, we start by reproducing the case studies presented in~\cite{bortolussi2016smoothed}, Network Epidemics and Prokaryotic Gene Expression, and then add a third biological model, Three-layer Phosphorelay. Finally, to have a better and unbiased quantification of the performances and of the scalability of SV-smMC, we test it over randomly generated pCTMC with parameter spaces of increasing dimensions. 
\begin{itemize}

    \item \textbf{Network Epidemics (SIR)}: a simple SIR model describing the spread of an epidemic in
a population of fixed size such that immunisation is permanent. 
The dynamics depends on two parameters $\beta$  and $\gamma$. The chosen STL property describes the termination of epidemics in a time between 100 and 120 time units from the epidemic onset: 
$\displaystyle \varphi =  (X_I > 0)\; \mathcal{U}_{[100,120]}\; (X_I = 0),$
where $X_I$ counts the infected nodes.

\item \textbf{Prokaryotic Gene Expression (PGE)}: model of LacZ, $X_{LacZ}$, protein synthesis in E. coli. The dynamics is governed by $11$ parameters $k_1$, $k_2$, $\dots$, $k_{11}$.
We choose an STL property for monitoring bursts of gene expression, rapid increases in LacZ counts followed by long periods of lack of protein production:
$\displaystyle \varphi = F_{[1600,2100]}(\Delta X_{LacZ} > 0 \wedge G_{[10,200]} (\Delta X_{LacZ} \le 0)),$
where $\Delta X_{LacZ}(t) = X_{LacZ}(t) - X_{LacZ}(t-1)$.

\item \textbf{Three-layer Phosphorelay (PR)}: network of three proteins $L1$, $L2$, $L3$ involved in a cascade of phosphorylation reactions (changing the state of the protein), in which  protein $Lj$, in its phosphorylated form $Ljp$, acts as a catalyser of phosphorylation of protein $L(j+1)$. There is a ligand $B$ triggering the first phosphorylation in the chain. The dynamics depends on $6$ parameters $k_p$, $k_1$, $k_2$, $k_3$, $k_4$, $k_d$.
The chosen STL property models a switch in the most expressed protein between $L1p$ and $L3p$ after time 300: 
$\displaystyle \varphi = G_{[0,300]}(L1p-L3p\ge0) \wedge F_{[300,600]}(L3p-L1p\ge0)$.

\item \textbf{Random pCTMC}: we randomly generate pCTMC over parameter spaces of increasing dimension. The number of interacting species $n$ and the type and the number of reactions are randomly determined. The STL properties considered captures different behaviours: $\varphi_1 = G_{[0,T]} (S_i \le S_j)$, $\varphi_2 = F_{[0,T]} (S_i \le S_j)$ and $\varphi_3 = F_{[0,T]} (\ G\ ( S_i < \tau)\ )$,
where $\tau$ and $T$ are two fixed hyper-parameters and the species to be monitored, $S_i$ and $S_j$, are randomly sampled for each property.  
\end{itemize}
Details about the dynamics, i.e. about the reactions, the selected initial states and the chosen parametric ranges, are provided in Appendix~\ref{app:casestudies}.

\subsection{Experimental Details}

\paragraph{Dataset generation.} The training set $D_t$ is built as per~\eqref{eq:train_dataset}. The test set, used to validate our results, can be summarized as $D_v = \big\{
    (\theta_j, (\ell_j^1,\dots ,\ell_j^{M_v}))\mid j=1, \dots,N_v
    \big\},  $ where $M_v$ is chosen very large, $M_v\gg M_t$, so that we have a good estimate of the true satisfaction probability over each test input. Input data, i.e. the parameter values, are scaled to the interval $[-1,1]$ to enhance the performances of the inferred models and to avoid sensitivity to different
scales in the parameter space.
In the first three case studies, the biology-inspired ones, we choose different subsets of varying parameters and train a separate model on each of these choices. In other words, we fix some of the parameters and let only the remaining ones vary. 
In particular, in SIR we consider the following configurations: $(a)$ fix $\gamma$ and let $\beta$ vary, $(b)$ fix $\beta$ and let $\gamma$ vary, $(c)$ vary both $\beta$ and $\gamma$; in PGE we consider the following configurations: $(d)$ $k_2$ is the only parameter allowed to vary, $(e)$ we let $k_2$ and $k_7$ vary; in PR we consider the following configurations: $(f)$ only $k_1$ varies, $(g)$ only $k_p, k_d$ vary, $(h)$ only $k_1,k_2,k_3$ vary, $(i)$ only $k_1,k_2,k_3,k_4$ vary, $(l)$ all six parameters are allowed to vary. 

On the other hand, to better analyze the scalability of SV-smMC we randomly generate pCTMC over parameter spaces $\Theta$ of dimension $2$, $3$, $4$, $8$, $12$, $16$ and $20$. For each dimension, we generate three different models and each model is tested over the three random properties $\varphi_1$, $\varphi_2$, $\varphi_3$ defined above. Table~\ref{tab:dataset_size_time} in Appendix~\ref{app:casestudies} shows the chosen dimensions for each generated dataset. In general, the number of observed parameters $N_t$ and the number of observations per point $M_t$ increase proportionally to the dimensionality of $\Theta$.

\paragraph{Experimental settings.} The CTMC dynamics is simulated via StochPy SSA\footnote{\url{https://github.com/SystemsBioinformatics/stochpy}} simulator for biology inspired models, whereas GillesPy2\footnote{\url{https://github.com/StochSS/GillesPy2}} is used to generate and simulate random pCTMC. The Boolean semantics of pcheck library\footnote{\url{https://github.com/simonesilvetti/pcheck}} is used to check the satisfaction of a certain formula for a specific trajectory. GPyTorch~\cite{gpytorch} library is used to train the SVI-GP models and Pyro~\cite{bingham2019pyro} library is used to train the SVI-BNN models, both built upon PyTorch~\cite{paszke2019pytorch} library. Instead, EP-GP is implemented in NumPy\footnote{Our implementation builds on \url{https://github.com/simonesilvetti/pyCheck/blob/master/smothed/smoothedMC.py}}. %
The experiments were conducted on a shared virtual machine with a 32-Core Processor, 64GB of RAM and an NVidia A100 GPU with 20GB, and 8 VCPU. 
Code and data will be presented for artifact evaluation. 

\paragraph{Training and evaluation.} We apply Stochastic Variational Inference on both Gaussian Processes (SVI-GPs) and Bayesian Neural Networks (SVI-BNNs) and compare them to the baseline smMC approach, where Gaussian Processes were inferred using Expectation Propagation (EP-GPs).
All models (EP-GP, SVI-GP and SVI-BNN) are Bayesian and trained over the training set $D_t$. 
Once the training phase is over, for each pair $\big(\theta_j, (\ell_j^1,\ldots, \ell_j^{M_v})\big)\in D_v$ in the test set, we obtain a probabilistic estimate of the satisfaction probability $f(\theta_j)$ (defined in Sect.~\ref{sec:sv-smc}). We compare such distribution
to the satisfaction probability $f_{\varphi}(\theta_j)$ estimated as the mean $\bar{L}_i$ over the Bernoulli trials $(\ell_j^1,\ldots, \ell_j^{M_v})$ and we call the latter SMC satisfaction probability.
We stress that SMC estimates provably converge to the true satisfaction probabilities, meaning that the width of confidence intervals converges to zero in the limit of infinite samples, while Bayesian inference quantifies the predictive uncertainty. Consequently, regardless of the number of samples, SMC and Bayesian estimates have different statistical meanings.

\paragraph{Evaluation metrics.} To define meaningful measures of performance, let's clarify the notation.
    For each point in the test set, $j\in\{1,\ldots,N_v\}$, let 
    $\bar{L}_j$ and $\sigma_j$ denote respectively the average and the standard deviation over the $M_v$ Bernoulli trials $(\ell_j^1,\ldots, \ell_j^{M_v})$. The inferred models, on the other hand, provide a posterior predictive distribution $p(f_j|\theta_j,D_t)$, 
let $q_{\epsilon}^j$ denote the $\epsilon$-th quantile of such distribution. 
The metrics used to quantify the overall performances of the models over each case study and each configuration are the following: 
\begin{enumerate}[(i)]
    \item the \emph{root mean squared error} (RMSE) between SMC and the expected satisfaction probabilities, i.e. the root of the average of the squared residuals
    $$
    \text{RMSE} = \frac{1}{N_v} \sum_{j=1}^{N_v} \big(\bar{L}_j-\mathbb{E}_{p(f|\theta_j,D_t)}[f(\theta_j)]\big)^2.
    $$
    This measure evaluates the quality of reconstruction provided by the mean of the posterior predictive distribution;

    \item the \emph{accuracy} over the test set, i.e. the fraction of non empty intersections between SMC confidence intervals and estimated $(1-\epsilon)$ credible intervals:
    \begin{align*}
    \text{Acc} = \frac{1}{N_v}\cdot
        \Bigg|\Bigg\{j\in\{1,\ldots,N_v\}:
            \Bigg[-\frac{z\,\sigma_j}{\sqrt{M_v}},\frac{z\,\sigma_j}{\sqrt{M_v}}\Bigg]\cap\Bigg[q_{\epsilon/2}^j,q_{1-\epsilon/2}^j\Bigg]\neq\emptyset
        \Bigg\}\Bigg|.\\
    \end{align*}
    In particular, we set $z=1.96$ and $\epsilon = 0.05$ in order to have the $95\%$ confidence intervals and the $95\%$ credible intervals respectively;
    \item the \emph{average width} of the estimated credible intervals
    $$\text{Unc} = \frac{1}{N_v} \sum_{j=1}^{N_v} \big(q_{1-\epsilon/2}^j-q_{\epsilon/2}^j\big),$$
    which quantifies how informative the predictive uncertainty is and allows us to detect over-conservative predictors.
\end{enumerate}
A good predictor should be balanced in terms of low RMSE, high test accuracy, i.e. high values for Acc, and narrow credible intervals, i.e. low values for Unc. 

\paragraph{Implementation details.} 
Both SVI-GP and SVI-BNN models are trained for $2k$ epochs with mini-batches of size $100$ and a learning rate of $0.001$. In SVI-GP the prior is computed on a maximum of $1k$ inducing points selected from the training set. SVI-BNNs have a fully connected architecture with $3$ layers and Leaky ReLU nonlinear activations. To evaluate SVI-BNNs we take $1k$ samples from the posterior distribution evaluated over test inputs. 

\paragraph{Prior tuning.} Choosing an adequate prior is of paramount importance. In this paper we leverage model-specific strategies to pick reasonable ones, however, no guarantees can be provided about the adequacy of such priors. 
For GP, the prior is strongly related to the chosen kernel and adapted to data by type II maximum likelihood maximization. In EP-GP the optimal kernel hyperparameters are searched beforehand (as in~\cite{bortolussi2016smoothed}), whereas in SVI-GP the kernel hyperparameters are optimized on the fly, i.e. the variational loss is maximized also w.r.t. the kernel hyperparameters. 
In SVI-BNN, a deterministic NN (with the same architecture of the BNN) is trained and the learned weights are used to center the prior distribution which is Gaussian with standard deviation equal to $1/m$ ($m$ is the layer-width).

\subsection{Experimental Results}

\paragraph{Computational costs.}
The cost of EP-GP inference is dominated by the cost of matrix inversion, which is cubic in the number of points in the training set. The cost of SVI-GP inference is cubic in the number of inducing points, which is chosen to be sufficiently small, and linear in the number of training instances. The cost of SVI-BNN is linear in the number of training points but it also depends on the architectural complexity of the chosen neural network. Variational models are trained by means of SGD, which is a stochastic inference approach. Thus, at least on simple configurations, it is likely to take longer than EP in reaching convergence. The computational advantage becomes significant as the complexity of the case study increases, i.e., when the training set is sufficiently large. EP faces memory limits, becoming unfeasible on configurations with parameter space of dimension higher than four. As a collateral advantage, SVI-GP optimizes the kernel hyperparameters on the fly during the training phase, whereas in EP-GP the hyperparameters search is performed beforehand and it is rather expensive.

\begin{figure*}[!t]
    \centering
    \includegraphics[scale=0.15]{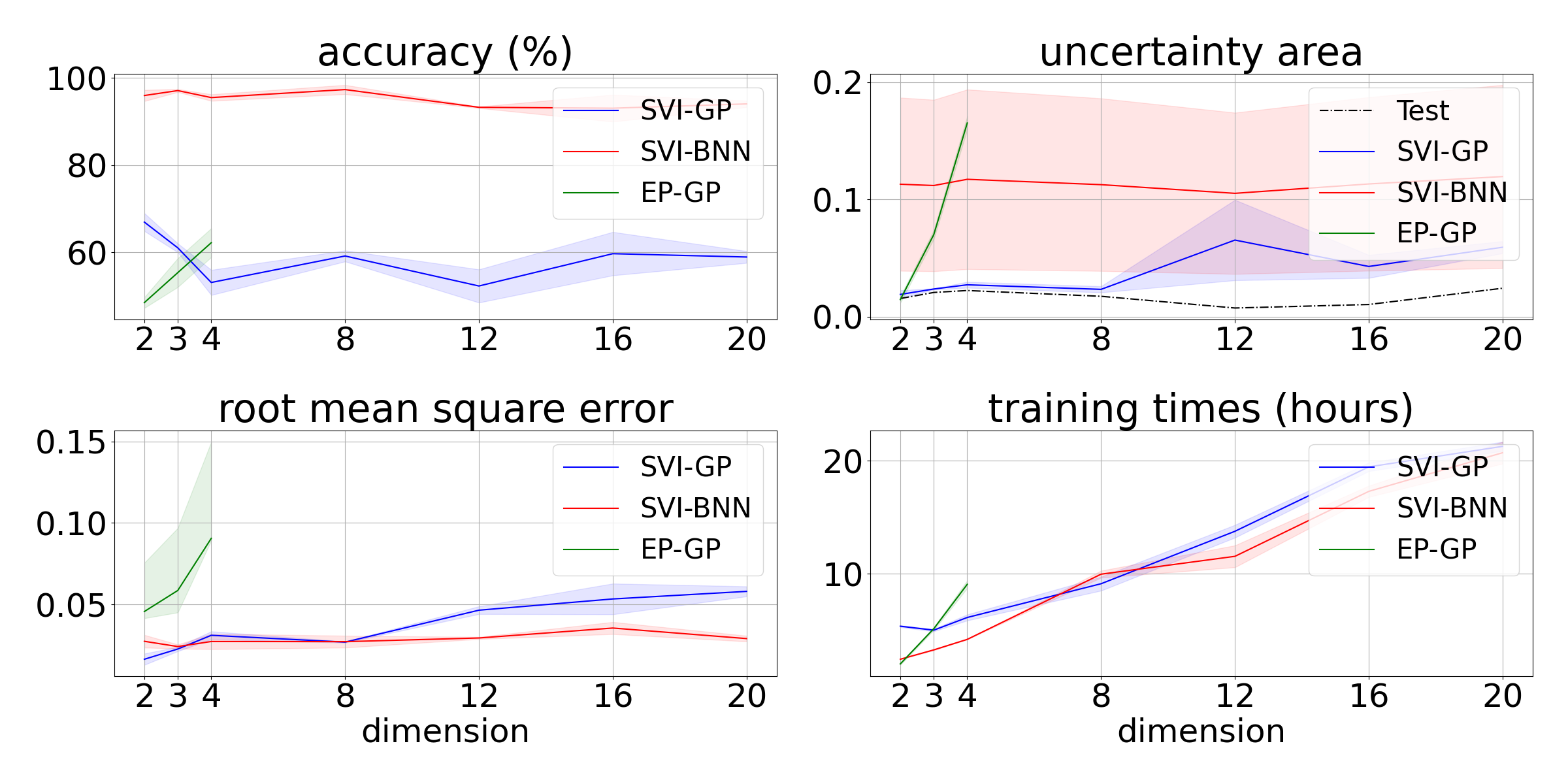}

\vspace{-0.2cm}
    
    \caption{Quantitative analysis of the scalability performances over randomly generated pCTMC with parameter spaces of increasing dimension (x-axis). For each dimension, we plot the $95\%$ confidence interval of Acc, RMSE and Unc over each test set corresponding to that dimension.\vspace{-0.5cm}}
    \label{fig:random_results}
\end{figure*}

For the randomly generated pCTMC we let the dimension of the training set grow linearly with the dimension of the parameter space ($N_t = 5000\cdot r$) and so does the training time of SVI-GP and SVI-BNN (see Fig.~\ref{fig:random_results}). On the other hand, Fig.~\ref{fig:random_results} shows how the training time of EP-GP grows much faster with respect to $r$ and becomes soon unfeasible ($r> 4$). SV-smMC is trained leveraging GPU acceleration. Its convergence times are comparable to EP's on simple configurations and they outperform EP on more complex ones.
Evaluation time for EP-GPs and SVI-GPs 
is negligible as it is computed from the analytic posterior. The evaluation time for SVI-BNNs  with $1k$ posterior samples is in turn negligible.

\vspace{-0.2cm}

\begin{table}[!b]
\centering
\adjustbox{max width=\textwidth, max height=\textheight}{
\begin{tabular}{l|c c c | c c c | c c c c }
    \toprule
    \textbf{Configuration} & \multicolumn{3}{c}{\textbf{RMSE}} & \multicolumn{3}{c}{\textbf{Accuracy}} & \multicolumn{4}{c}{\textbf{Uncertainty}}\\
    & EP-GP & SVI-GP & SVI-BNN &  EP-GP & SVI-GP & SVI-BNN & Test &  EP-GP & SVI-GP & SVI-BNN \\
    \hline
    $(a)$ SIR $\beta$ & $1.41$ & $\mathbf{1.38}$ & $1.43$ & $\mathbf{100.00}$ & $98.80$ & $\mathbf{100.00}$ & $0.044$ & $0.044$ & $\color{teal}\underline{ 0.032}$ & $\color{purple}\overline{ 0.097}$\\
    $(b)$ SIR $\gamma$ & $1.02$ & $\mathbf{0.84}$ & $0.96$& $77.80$ & $77.30$ & $\mathbf{92.60}$& $0.018$ & $0.018$ & $\color{teal}\underline{ 0.012}$ & $\color{purple}\overline{0.058}$\\
    $(c)$ SIR $\beta,\gamma$ & $1.25$ & $1.20$ &$\mathbf{0.99}$& $72.75$ & $85.75$ & $\mathbf{92.25}$& $\color{teal}\underline{0.019}$ & $0.042$ & $\color{teal}\underline{0.019}$ & $\color{purple}\overline{0.047}$ \\
    \hline
    $(d)$ PGE $k_2$ & $\mathbf{5.34}$ & $5.83$ & $5.51$ & $93.75$ & $94.00$ & $\mathbf{97.50}$& $0.039$ & $\color{teal}\underline{0.030}$ & $0.037$ & $\color{purple}\overline{0.080}$\\
    $(e)$ PGE $k_2, k_7$ & $6.48$ & $3.44$ & $\mathbf{2.06}$ & $78.25$ & $89.25$ & $\mathbf{95.00}$& $0.043$ & $\color{teal}\underline{0.021}$ & $0.056$ & $\color{purple}\overline{0.093}$\\
    \hline
    $(f)$ PR $k_1$ & $2.25$ & $1.99$ & $\mathbf{1.78}$& $99.70$ & $99.20$ & $\mathbf{100.00}$& $0.058$ & $0.059$ & $\color{teal}\underline{0.046}$ & $\color{purple}\overline{0.100}$\\
    $(g)$ PR $k_p,k_d$ & $2.95$ & $2.31$ & $\mathbf{1.89}$ & $99.25$ & $96.25$ & $\mathbf{99.75}$& $0.055$ & $\color{purple}\overline{0.108}$ & $\color{teal}\underline{0.037}$ & $0.093$\\
    $(h)$ PR $k_1,k_2,k_3$ & $6.97$ & $2.30$ & $\mathbf{2.01}$  & $99.80$ & $93.70$ & $\mathbf{100.0}$ & $0.050$ & $\color{purple}\overline{0.340}$ & $\color{teal}\underline{0.030}$ & $0.121$ \\
    $(i)$ PR $k_1,\ldots,k_4$ &  $10.63$ & $\mathbf{2.44}$ & $2.67$& $99.02$ & $93.65$ & $\mathbf{99.92}$& $0.050$ & $\color{purple}\overline{0.682}$ & $\color{teal}\underline{0.030}$ & $0.150$\\
    $(l)$ PR $k_p,k_1,\ldots,k_4,k_d$ & 
    - & $1.87$ & $\mathbf{1.56}$  & - & $97.02$ & $\mathbf{99.80}$ & 0.049 & - & $\color{teal}\underline{0.028}$ & $\color{purple}\overline{0.087}$\\
    
    \bottomrule
\end{tabular}}
\caption{Root Mean Square Error ($\times10^{-2}$), test accuracy ($\%$) and average uncertainty width for EP-GP, SVI-GP and SVI-BNN. SVI-BNNs are evaluated on $1k$ posterior samples. For each case study, we highlight the minimum MSE, the highest accuracy values and the \textcolor{teal}{lowest} and \textcolor{purple}{highest} uncertainty values. Uncertainty is compapurple to the average uncertainty width of the test set.\vspace{-0.5cm}}
\label{tab:mse_acc_unc}
\end{table}

\paragraph{Performance evaluation.} The evaluation metrics are the root mean square error (RMSE), the accuracy (Acc) and the width of the uncertainty quantification area (Unc). Results over the randomly generated pCTMC are summarized in Fig.~\ref{fig:random_results}, whereas results over the biological case studies are summarized in Table~\ref{tab:mse_acc_unc}. Fig.~\ref{fig:random_results} shows, for each dimension, the $95\%$ confidence interval of Acc, RMSE and Unc over each test set corresponding to that dimension. More precisely, each dimension has nine associated datasets, three models with three properties each, so that nine different smMC models have been trained. Notice how the width of the inferred credible intervals is compared against the width of SMC confidence intervals.

In addition, Fig.~\ref{fig:SIR_beta}-Fig.~\ref{fig:PhosRelay_k1} in Appendix~\ref{app:plots} show the results over one-dimensional configurations - $(a)$, $(b)$, $(d)$ and $(f)$ respectively - whose results over the test set are easy to visualise. In particular, we show the mean and the $95\%$ credible intervals of the estimated satisfaction probability $f(\theta_j)$ for EP-GPs, SVI-GPs and SVI-BNNs.
Fig.~\ref{fig:SIR_betagamma}-Fig.~\ref{fig:PhosRelay_kpkd} in Appendix~\ref{app:plots} compare the results of EP-GPs, SVI-GPs and SVI-BNNs over two-dimensional configurations - $(c)$, $(e)$ and $(g)$. In particular, we compare the SMC estimate of the satisfaction probability $f_{\varphi}(\theta_j)$ to the average satisfaction probability $\mathbb{E}[f(\theta_j)]$ estimated by EP-GPs, SVI-GPs and SVI-BNNs over each input $\theta_j$ of the test set.

We now compare the performances obtained by the variational approaches of SV-smMC to those of the smMC baseline based on EP-GP. Table~\ref{tab:mse_acc_unc} and Fig.~\ref{fig:random_results} show how the RMSE of SV-smMC solutions is almost always lower than that of smMC. 
In addition, the baseline solution presents an RMSE that grows proportionally to the complexity and the dimensionality of the underlying configuration. On the contrary, SV-smMC solutions do not reflect such behaviour, as the value of the RMSE is almost constant across all the different configurations. 
About the informativeness of uncertainty estimations, we notice how SVI-BNN tends to produce credible intervals that are always larger than the one of SVI-GP, which, in turn, tends to underestimate the underlying uncertainty. This phenomenon appears in all the different configurations and it is easily observable in Fig.~\ref{fig:random_results}. We argue that SVI-GP tends to provide overconfident predictions due to the sparsification strategies used during inference. Such behaviour is well-known and discussed in~\cite{Candela2005AUV,Rasmussen2005HealingTR}.
SVI-BNN does not present such behaviour as it does not undergo any sparsification. On the other hand, the baseline smMC tends to have tight uncertainty estimates on low-dimensional configurations, but it becomes excessively over-conservative in high-dimensional configurations, making the predicted credible intervals almost uninformative.
All models reach extremely high accuracies over the test set. 
SVI-BNN reaches the best performances over all the configurations: the average accuracy is around $95 \%$ and it is always higher than $71\%$. This result is not surprising given that SVI-BNN shows low RMSEs (overall average around $0.02$) and slightly over-conservative credible intervals (overall average uncertainty width around $0.1$). The SVI-GP accuracy, on the other hand, fluctuates around $60\%$ as it tends to provide over-confident credible intervals with an overall average RMSE comparable to that of SVI-BNN but with overall average uncertainty width of around $0.03$.

\begin{figure}[!t]
    \centering
    \includegraphics[scale=0.12]{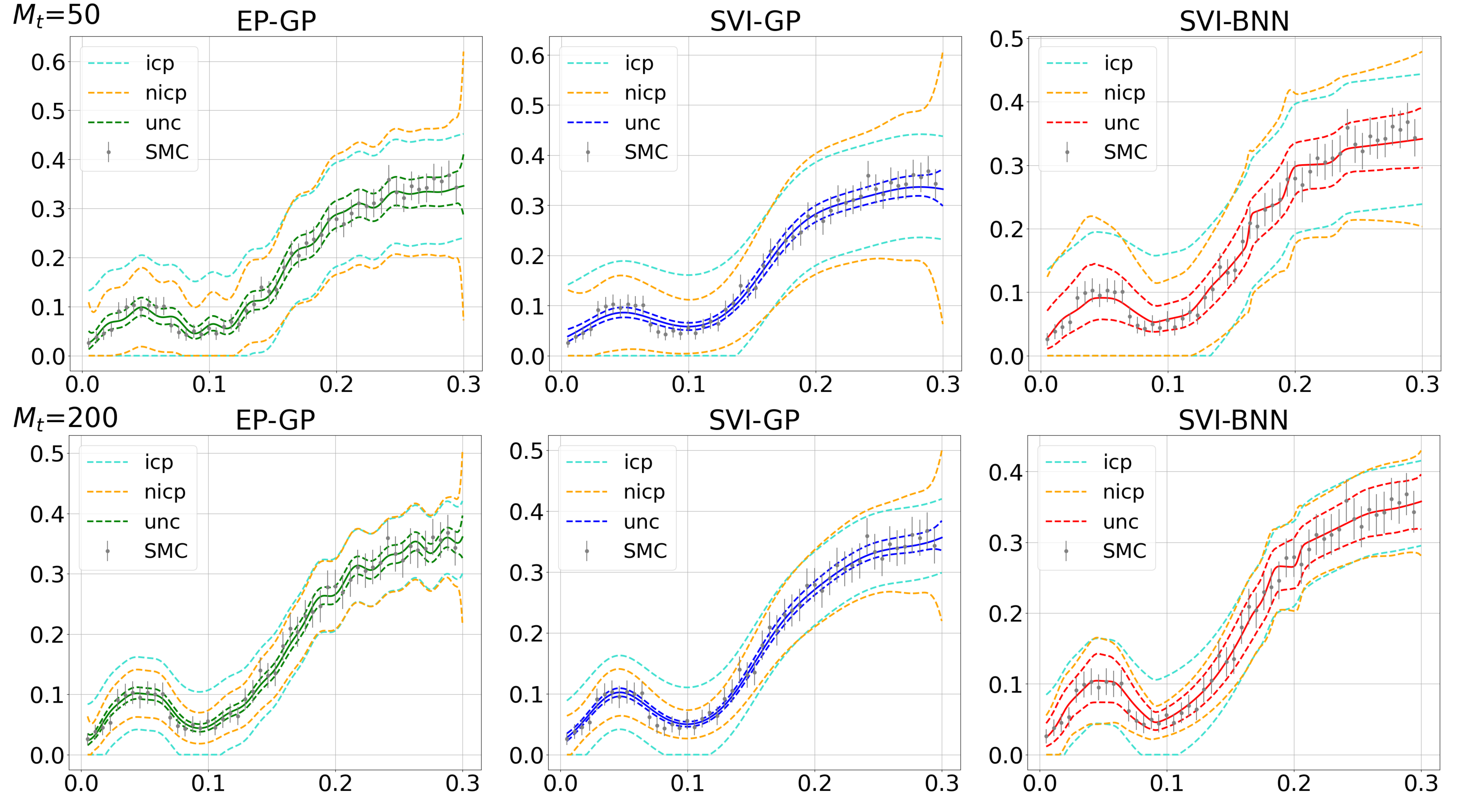}

\vspace{-0.2cm}
    
    \caption{Comparison of the point-wise Bayesian estimate of uncertainty with those of ICP (cyan) and of the normalized ICP (orange) with guaranteed coverage of $95\%$ for increasing values of $M_t$. \vspace{-0.5cm}}
    \label{fig:unc_comp_1}
\end{figure}

\paragraph{Statistical guarantees.} We compare the quantification of predictive uncertainty provided by the Bayesian SV-smMC with those of ICP and NICP. For ease of discussion, we focus on configuration $(a)$, set $N_t=500$, $N_c= 200$ and show how the performances vary for an increasing number of observations $M_t$. The confidence level is set to $\varepsilon=0.05$. Fig.~\ref{fig:unc_comp_1} compares the point-wise bounds around the predictive mean as a function of the one-dimensional $\Theta$.
Fig.~\ref{fig:avg_unc_500} in Appendix~\ref{app:plots} shows how the average widths (over $D_v$) of the bounds vary in function of the number $M_t$ of observations per point.
The tightness of the bounds depends on the number $N_t$ of observed parameters, on the number $M_t$ of Bernoulli observations per point, which controls the level of noise present in the data, and on the quality of the inferred model $q(f)$. 
Given that the guarantees provided by ICP and NICP are over an SMC estimate of $f_\varphi$, one should correct the error estimates leveraging Chernoff bounds. For instance, if $f_\varphi$ is estimated using $M_t = 500$ observations per point, the bounds should be expanded by $\sim 0.06$ in order to have a guaranteed coverage of $0.9$ of the exact satisfaction function $f_\varphi$. See Appendix~\ref{app:icp} for the derivation of this Chernoff bound over SMC estimates of $f_\varphi$ and Fig.~\ref{fig:unc_comp_2} in the same Appendix to see how the width of Chernoff bounds changes as $M_t$ increases. 
In general, the widths of all bounds decrease as $M_t$ increases. 
Despite the nice rigorous guarantees provided by ICP and NICP, we notice how these bounds are less tight compared to the Bayesian counterpart that, on the other hand, provides no coverage guarantees. We can also notice how the bounds of  BNN tend to be closer to NICP ones for 200 (and 500) observations per point and how NICP corrects overconfident uncertainty estimates of sparsified GP. 
Let us stress how the Bayesian quantification of uncertainty is a key ingredient for Theorem~\ref{thm:nicp} to hold. If one used a deterministic predictor to approximate the satisfaction function we would not be able to leverage the NICP framework, and have only a looser uniform bound. 

For completeness, we also looked at the statistical guarantees provided by PAC Bayes theory. In particular, we adapted Catoni’s bound~\cite{Catoni2004APA} to our smMC framework. 
The resulting bounds are typically much broader compared to ICP and NICP and hold only in expectation over $\Theta$, thus are less effective in practice. 
Theory and results are in Appendix~\ref{app:pac}. 

\paragraph{Discussion.} To summarize, we can see how, in general, SV-smMC solutions scale a lot better to high-dimensional problems compared to smMC, both in terms of feasibility and in terms of quality of the results. SVI-BNN reaches the highest accuracy and provides rather conservative predictions. SVI-GP, on the other, reaches low RMSEs and tends to provide overconfident predictions. Finally, we see how EP-GP is competitive only on extremely simple configurations. As the dimensionality increases, so does the error: the RMSE increases and the credible intervals become excessively broad. Moreover, we soon reach the memory-bound wall that makes EP-GP solution unfeasible on configurations with more than four parameters.

\section{Conclusions}\label{sec:conclusions}

This paper presents SV-smMC, an extension of Smoothed Model Checking, based on stochastic variational inference, that scales well to high dimensional parameter spaces and that enables GPU acceleration.
\new{We further enrich smMC and SV-smMC with statistical guarantees on the error committed, based on an integration of uncertainty estimates of Bayesian methods within the framework of Inductive Conformal Predictions. This method is cheap to compute and allows us even to correct for overconfident uncertainty estimates of sparsified Gaussian Processes.   }
In addition, this paper offers a comparison of the performances of stochastic variational inference over two different Bayesian approaches - namely Gaussian processes (SVI-GP) and Bayesian neural networks (SVI-BNN) - against those of the baseline smMC, based on the expectation propagation technique.
In particular, our experiments show that the posterior predictive distribution provided by SVI-BNN provides the best overall results in terms of the estimated satisfaction probabilities. On the other hand, thanks to GPU acceleration, SVI-GP is able to achieve competitive performances with a significant speed-up in computational time. Furthermore, we 
show how variational approaches are able to overcome the computational limitations of the expectation propagation algorithm over large datasets. 

SV-smMC can be naturally extended with active learning ideas, following the line of~\cite{TACAS18,TCS15,FM17}, solving efficiently parameter synthesis and design tasks.

\bibliographystyle{splncs04}
\bibliography{references}

\clearpage
\appendix

\section{Stochastic Variational Inference over GP and BNN}\label{app:svi_details}

\subsection{SVI over Gaussian Processes}\label{app:svi_gp}

A \textit{Gaussian Process (GP)} is a stochastic process, i.e., a collection of random variables indexed by some input variable, in our case $\theta\in\Theta$, such that every finite linear combination of them is normally distributed.
In practice, a GP defines a distribution over real-valued functions of the form $g: \Theta\rightarrow\mathbb{R}$ and such distribution is uniquely identified by its mean and covariance functions, respectively denoted by $\mu(\theta)=\mathbb{E}[g(\theta)]$ and $k(\theta,\theta')$. The GP can thus be denoted as $\mathcal{GP}(\mu(\theta),k(\theta,\theta'))$. This means that the function value at any point $\theta$, $g(\theta)$, is a {Gaussian} random variable with mean $\mu(\theta)$ {and variance $k(\theta,\theta)$}.  Typically, the covariance function $k(\cdot, \cdot)$ depends on some hyper-parameter $\gamma$. 

GPs model the posterior probabilities by defining latent functions $g:\Theta\to \mathbb{R}$, whose output values are then mapped into the $[0, 1]$ interval
by means of a so-called link function $\Phi$, typically the logit or the probit function. 

Given an input $\theta_i$, let $g_i = g(\theta_i)$ denote its latent variable, i.e., the latent function $g$ evaluated at $\theta_i$. Also denote $g_t = [ g(\theta_i) | \theta_i \in \Theta_t ]$, where $\Theta_t$ is the set of input points of the training set $D_t$. 
From $\Theta_t$ it is possible to compute the mean vector $\mu_t$ of the GP, by evaluating the mean function $\mu(\cdot)$ at every point in the set, and the covariance matrix $K_{N_tN_t}$, by evaluating the covariance function on every pair of points in the set: $\mu_t = [\mu(\theta_i) | \theta_i\in \Theta_t]$ and $K_{N_tN_t}=[ k_{\gamma}(\theta_i,\theta_j)|\theta_i,\theta_j\in \theta_t]$. 

The first step of a GP algorithm is to place a GP prior over the latent function $g$, defined by 
$$p(g|\Theta_t) = \mathcal{N}(g|\mu_t, K_{N_tN_t}).$$

Let now consider a test input $\theta_*$ with latent variable $g_*$. In order to do inference, that is, predict its label  $\ell_*$, we have to compute 
\begin{equation}\label{eq:pred_gp_app}
    p(\ell_*|\theta_*,\Theta_t,L_t) = \int \Phi (g_*)p(g_*|\theta_*, \Theta_t,L_t)dg_*,
\end{equation}
where $L_t$ denotes the set of Boolean tuples corresponding to points in $\Theta_t$.

To compute equation~\eqref{eq:pred_gp_app}, we have to marginalize the posterior over the latent Gaussian variables:
\begin{equation}\label{eq:post_gp_app}
    p(g_*|\theta_*, \Theta_t,L_t) = \int p({g}_*|\theta_*, \Theta_t,g_t)p(g_t|\Theta_t,L_t)d g_t,
\end{equation}
where the posterior $p(g_t|\Theta_t,L_t)$ can be obtained using the standard Bayes rule
\[
p(g_t|\Theta_t,L_t)=\frac{p(L_t|g_t,\Theta_t)p(g_t|\Theta_t)}{p(L_t|\Theta_t)}.\]
Therefore, performing inference reduces to solving two integrals, \eqref{eq:pred_gp_app} and \eqref{eq:post_gp_app}. In classification, the first integral is not available in closed form since it is the convolution of a Gaussian distribution, $p(g_t|\Theta_t)$, and a non-Gaussian one, $p(L_t|g_t,\Theta_t)$ (binomial). Hence, we have to rely on approximations in order to compute and integrate over the posterior $p(g_t|D_t)$. 
In our experiments, we use stochastic variational inference which provides a Gaussian approximation $q(g_t|D_t)$ of the posterior $p(g_t|D_t)$, which can then be easily computed and integrated over.

\paragraph{Stochastic Variational Inference.}
Here we describe how SVI works over GP with non-Gaussian likelihoods. For a more detailed description, we refer the interested reader to~\cite{hensman2013gaussian,hensman2015scalable}. 
Variational approaches to GP start with sparsification, i.e. defining a set of  $m \ll N_t$ inducing points $Z = \{z_1,\dots,z_m\}$ that live in the same space of $\Theta_t$ and, from them, we define a set of inducing variables $u_t = [g(z_i)|z_i\in Z]$. The covariance matrix over inducing points, $K_{mm}$, is less expensive to invert and thus it act as a low-rank approximation of $K_{N_tN_t}$.
The goal of the sparse SVI solution is to propose a parametric distribution $q(u_t)$ over inducing variables $u_t$ that minimizes the approximate posterior distribution $p(u_t|D_t))$, i.e. $KL(q(u_t)||p(u_t|D_t))$.
Such KL divergence can be written as
\begin{align*}
  KL&(q(u_t)||p(u_t|D_t)) = \mathbb{E}_{q(u_t)} \left[ \log \frac{q(u_t)}{p(D_t|u_t)p(u_t)}p(D_t)\right] = \\
  &= \mathbb{E}_{q(u_t)} \left[ \log q(u_t)-\log p(D_t|u_t) - \log p(u_t)\right]+\log p(D_t) = \\
  &= KL(q(u_t)||p(u_t))- \mathbb{E}_{q(u_t)}\left[\log p(D_t|u_t)\right]+\log p(D_t).
\end{align*}
Since the KL diverge is always positive, we can rewrite the above identity as a lower bound for the marginal log likelihood:
\begin{equation}\label{eq:kl_lb_gp}
 \log p(D_t)\ge   \mathbb{E}_{q(u_t)}\left[\log p(D_t|u_t)\right]-KL(q(u_t)||p(u_t)),
\end{equation}
where the first term is the expected log-likelihood of our data with respect to values of $u_t$ sampled from $q(u_t)$, whereas the second term denotes the Kullback-Leibler divergence~\cite{joyce2011kullback} between $q(u_t)$ and $p(u_t)$. 
Leveraging Jensen inequality we know that
\begin{equation}\label{eq:cond_JI}
    \log p(D_t|u_t)\ge \mathbb{E}_{p(g_t|u_t)}\left[\log p(D_t|g_T)\right].
\end{equation}
By substituting~\eqref{eq:cond_JI} in~\eqref{eq:kl_lb_gp} and defining $q(g_t) := \int p(g_t|u_t)q(u_t)du_t$, we obtain the following lower bound for the marginal log likelihood:
\begin{equation}
 \log p(D_t)\ge   \mathbb{E}_{q(g_t)}\left[\log p(D_t|g_t)\right]-KL(q(u_t)||p(u_t)),
\end{equation}
where the first term is now the expected log-likelihood of our data with respect to values of $g_t$ sampled from $q(g_t)$.

We choose $q(u_t) := \mathcal{N}(u_t|\eta, \tilde{S})$, where $\tilde{S}$ is defined using a lower triangular form, $\tilde{S} := SS^T$, in order to maintain positive-definiteness. Then, with some algebraic manipulation, we express the variational distribution over $g_t$ as
\begin{equation}\label{eq:gp_qg}
    q(g_t) = \mathcal{N}(g_t|A\eta, K_{N_tN_t}+A(\tilde{S}-K_{mm})A^T),
\end{equation}
where $A = K_{N_tN_t}K_{mm}^{-1}$. Since our
likelihood factors as $p(D_t | g_t ) = \prod_{i=1}^{N_t} p(L_i | g_i )$, the lower bound becomes:
\begin{equation}\label{eq:gp_svi_bound_app}
   \log p(D_t)\ge \sum_{i=1}^{N_t}\mathbb{E}_{q(g_i)}[\log p(L_i|g_i)]-KL[q(u_t)||p(u_t)]:= \mathcal{L}_{GP}(\eta, S, Z, \gamma).
\end{equation}
The SVI algorithm then consists of maximizing $\mathcal{L}_{GP}$ with respect to its parameters using  gradient-based stochastic optimization. The gradient of $\mathcal{L}_{GP}$ contains the partial derivatives w.r.t. the SVI hyper-parameters and w.r.t. the hyper-parameter $\gamma$ of the covariance function. Computing the KL divergence in~\eqref{eq:gp_svi_bound_app} requires only $\mathcal{O}(m^3)$ computations. Most of the work will thus be in computing the
expected likelihood terms. 
Given the ease of parallelizing
the simple sum over $N_t$, we can optimize $\mathcal{L}_{GP}$ in
a distributed or in a stochastic fashion by selecting mini-batches of the data at random.

\paragraph{Predictive distribution.} The posterior of \eqref{eq:post_gp_app} is now approximated as $p(g_*|L_t)\approx \int p(g_*|u_t)q(u_t)du_t: = q(g_*)$. The integral above can be treated similarly to~\eqref{eq:gp_qg}, computing its mean and variance takes $\mathcal{O}(m^2)$. From the mean and the variance of $q(g_*)$, we obtain the respective credible interval and use the link function $\Phi$ to map it to a subset of the interval $[0,1]$, so that we have mean and credible interval of the posterior predictive distribution $p(f_*|\theta_*, D_t)$.

\subsection{SVI over Bayesian Neural Networks}\label{app:svi_bnn}

The core idea of Bayesian neural networks (BNNs) is to place a
probability distribution over the weights of a neural network, transforming the latter into a probabilistic model.

The Bayesian learning process starts by defining a prior distribution $p(\mathbf{w})$ for $\mathbf{w}$ that expresses our initial belief about the parameter values. A common choice is to choose a zero-mean Gaussian prior. As we observe data $D_t$, we update this prior to a posterior distribution using Bayes' rule:
\begin{equation}\label{eq:bnn_post_app}
    p(\mathbf{w}| D_t) = \frac{p(D_t| \mathbf{w})p(\mathbf{w})}{p(D_t)}.
\end{equation}

Similarly to GP, the likelihood $p(L_i| \theta_i,\mathbf{w})$ is the Binomial distribution 
\begin{equation}\label{eq:bnn_bin_lkh_app}
p(L_i| \theta_i,\mathbf{w}) = 
Binomial \left(L_i|  M_t, f_{\mathbf{w}}(\theta_i))
\right).
\end{equation}
It follows that, given a set of i.i.d.\ observations $D_t$, the likelihood function can be expressed as 
\begin{equation}
    p(D_t| \mathbf{w}) = \prod_{(\theta_i,L_i)\in D_t} p(L_i| \theta_i,\mathbf{w}).
\end{equation}

Note that, because of the non-linearity introduced by the
neural network function $f_{\mathbf{w}}(\theta)$ and by the Binomial likelihood, the posterior $p(\mathbf{w}|D_t)$ is non-Gaussian. 
Finally, in order to predict the value of the target for an unobserved input $\theta_*$, we marginalize
the predictions with respect to the posterior distribution
of the parameters, obtaining 
\begin{equation}\label{eq:bnn_predictive_app}
    p(f_*|\theta_*, D_t) = \int p(f_*|\theta_*, \mathbf{w}) p(\mathbf{w}|D_t)d\mathbf{w}.
\end{equation}

The latter is called \emph{posterior predictive} distribution and it can be used to retrieve information about the uncertainty of a specific prediction $f_* := f_\mathbf{w}(\theta_*)$.
Unfortunately, the integration is analytically intractable due to the non linearity of the neural network function~\cite{bishop2006pattern,mackay1992practical}.

\paragraph{Stochastic Variational Inference}
Since precise inference is infeasible, we consider an approximate inference method based on Stochastic Variational Inference (SVI)~\cite{jordan1999introduction}.
SVI directly approximates the posterior distribution with a known parametric distribution $q_\psi(\mathbf{w})$, typically a distribution easy to sample from. Its parameters $\psi$, called variational parameters, are learned by minimizing the Kullback-Leibler (KL) divergence between the proposed distribution and the posterior. The KL divergence between $q_\psi(\mathbf{w})$ and $p(\mathbf{w}|D_t)$ is defined as
\begin{equation}
    KL(q_\psi(\mathbf{w})||p(\mathbf{w}|D_t) ) = \int q_\psi(\mathbf{w})\log\frac{q_\psi(\mathbf{w})}{p(\mathbf{w}|D_t)}d\mathbf{w}.
\end{equation}
The goal is thus to find the members of the variational family that are closest to the true posterior $p(\mathbf{w}|D_t)$.
Since the posterior distribution is not known, we need to find an objective function that does not depend explicitly on the unknown posterior.

In order to do so, we rewrite the KL objective as follows:
\begin{align*}
    KL&(q_\psi(\mathbf{w})||p(\mathbf{w}|D_t)) = \mathbb{E}_{q_\psi(\mathbf{w})}\left[\log\frac{q_\psi(\mathbf{w})}{p(D_t|\mathbf{w})p(\mathbf{w})}p(D_t)\right]=\\
    &=  \mathbb{E}_{q_\psi(\mathbf{w})}\left[\log q_\psi(\mathbf{w}) - \log p(D_t|\mathbf{w}) -\log p(\mathbf{w})\right]+\log p(D_t)=\\
    &= KL(q_\psi(\mathbf{w})||p(\mathbf{w})) - \mathbb{E}_{q_\psi(\mathbf{w})}\log p(D_t|\mathbf{w})+\log p(D_t).
\end{align*}

As the KL divergence is always positive, we can rephrase the above equivalence as a lower bound for the log marginal likelihood:
\begin{equation}\label{eq:evidence_lb}
    \log p(D_t) \ge \mathbb{E}_{q_\psi(\mathbf{w})}\log p(D_t|\mathbf{w}) - KL(q_\psi(\mathbf{w})||p(\mathbf{w})):=\mathcal{L}_{BNN}(\psi).
\end{equation}

This lower bound, called Evidence Lower Bound (ELBO)~\cite{deodato2019bayesian}, can be used as a new objective function.
The first term is the expected log-likelihood of our data with respect to values of $f_\mathbf{w}$ sampled from $q_\psi(\mathbf{w}|D_t)$, whereas the second term is the KL divergence between the proposal distribution and the prior. 
The family of distribution $q_\psi$ should be a distribution easy to sample from and such that the KL divergence is easy to compute. A common choice for $q_\psi$ is the Gaussian distribution (where $\psi$ denotes its mean and variance). The KL distribution among two Gaussian distributions has an exact analytical form. Thus, the ELBO of~\eqref{eq:evidence_lb} can be computed and it can be used as the objective function of a maximization problem over $\psi$.
In SVI, the variational objective, i.e., the negative
ELBO, becomes the loss function used to train a Bayesian neural network~\cite{deodato2019bayesian}. 

\paragraph{Empirical approximation of the predictive distribution.} The predictive distribution~\eqref{eq:bnn_predictive_app} is a non-linear combination of Gaussian distributions, and thus it is not Gaussian. However, samples can be easily extracted from $q_\psi(\mathbf{w})$, which allows us to obtain an empirical approximation of the predictive distribution. Let $\it [w_1,\dots, w_C]$ denote a vector of $C$ realizations of the random variable $\mathbf{w}\sim q_\psi(\mathbf{w})$. Each realization $w_i$ induces a deterministic function $f_{w_i}$ that can be evaluated at $\theta_*$, the unobserved input, providing an empirical approximation of $p(f_*|\theta_*, D_t)$. By the strong law of large numbers, the empirical approximation converges to the true distribution as $C\to \infty$~\cite{van2000asymptotic}. The sample size $C$ can be chosen, for instance, to ensure a given width of the confidence interval for a statistic of interest~\cite{rasch2011optimal} or to bound the probability that the empirical distribution differs from the true one by at most some given constant~\cite{massart1990tight}.

\section{Case Studies}\label{app:casestudies}

\subsection{Biological Case Studies}
\begin{itemize}

    \item \textbf{Network Epidemics (SIR)}: spread of an epidemics in
a population of fixed size. Here we consider the case of permanent immunisation.

Reactions:
\begin{itemize}
    \item[] $R_1: S+I \to I+I$ with rate function $k_i X_SX_I$,
    \item[] $R_2: I \to R$ with rate function $k_r X_I$.
\end{itemize}

Initial state: $X_S = 95$, $X_I = 5$, $X_R = 0$ and $N = 100$ ($\Delta t = 0.5$). 
In configuration $(a)$, $\beta$ varies in the interval $[0.005,0.3]$ and $\gamma = 0.05$; in $(b)$, $\beta=0.12$ and $\gamma$ varies in $[0.005,0.2]$; whereas in $(c)$, $\beta$ varies in the interval $[0.005,0.3]$ and $\gamma$ varies in $[0.005,0.2]$.

\item \textbf{Prokaryotic Gene Expression}: a more complex model that captures LacZ protein synthesis in E. coli. 

Mass-action reactions:
\begin{itemize}
    \item[] $R_1: PLac + RNAP \to PLacRNAP$ with rate $k_1$,
    \item[] $R_2: PLacRNAP \to PLac+RNAP$ with rate $k_2$,
    \item[] $R_3: PLacRNAP \to TrLacZ1$ with rate $k_3$,
    \item[] $R_4: TrLacZ1 \to RbsLacZ + PLac + TrLacZ2$ with rate $k_4$,
    \item[] $R_5: TrLacZ2 \to RNAP$ with rate $k_5$,
    \item[] $R_6: Ribosome + RbsLacZ \to RbsRibosome$ with rate $k_6$,
    \item[] $R_7: RbsRibosome \to Ribosome + RbsLacZ$ with rate $k_7$,
    \item[] $R_8: RbsRibosome \to TrRbsLacZ+RbsLacZ$ with rate $k_8$,
    \item[] $R_9: TrRbsLacZ \to LacZ$ with rate $k_9$,
    \item[] $R_{10}:PLacZ \to dgrLacZ$ with rate $k_{10}$,
    \item[] $R_{11}: RbsLacZ \to dgrRbsLacZ$ with rate $k_{11}$.
\end{itemize}
Initial state: $PLac = 1$, $RNAP = 35$, $Ribosome = 350$, $PLacRNAP = TrLacZ1 = RbsLacZ = TrLacZ2 = RbsRibosome = TrRbsLacZ = LacZ = dgrLacZ = dgrRbsLacZ = 0$.

The default parametric values are
$k_1 = 0.17$,
$k_2 = 10$,
$k3=k_4 = 1$,
$k_5 = k_9 = 0.015$,
$k_6 = 0.17$,
$k_7 = 0.45$,
$k_8 = 0.4$,
$k_{10} = 0.0000642$,
$k_{11} = 0.3$.
In configuration $(d)$, $k_2$ varies in $[10,100000]$, whereas in configuration $(e)$, $k_2$ varies in $[10,10000]$ and $k_7$ varies in $[0.45,4500]$.

\item \textbf{Three-layer Phosphorelay}: network of three layers $L1$, $L2$, $L3$. Each layer can be found also in phosphorylate form $L1p$, $L2p$, $L3p$ and there is a ligand $B$. Initial state: $L1p=L2p=L3p = L4p = B = 0$, $L1=L2=L3 = 32$ and $N = 5000$.

Reactions:
\begin{itemize}
    \item[] $R_1: \emptyset \to B$ with rate function $k_p$
    \item[] $R_2: L1+B\to L1p+B$ with rate function $k_1\cdot L1\cdot B/N$
    \item[] $R_3: L1p+L2\to L1+L2p$ with rate function $k_2\cdot L1p\cdot L2/N$
    \item[] $R_4: L2p+L3\to L2+L3p$ with rate function $k_3\cdot L2p\cdot L3/N$
    \item[] $R_5: L3p\to L3$ with rate function $k_4\cdot L2p/N$
    \item[] $R_6: B \to \emptyset$ with rate function $k_d\cdot B$
\end{itemize}

\end{itemize}

The default parametric values are $k_p = 0,1$, $k_d = 0.05$, $k_1=k_2=k_3 = 1$ and $k_4 = 2$.
In configuration $(f)$, $k_1$ varies in $[0.1,2]$; in $(g)$, $k_p$ varies in $[0.01,0.2]$ and $k_d$ varies in $[0.005,0.1]$; in $(h)$, $k_1, k_2$ and $k_3$ all vary in the interval $[0.1,2]$; in $(i)$, $k_1, k_2, k_3$ vary in the interval $[0.1,2]$ and $k_4$ varies in $[0.5,5]$; whereas in configuration $(l)$, $k_1, k_2, k_3, k_4$ vary as in $(i)$ and $k_p, k_d$ varies as in $(g)$.

\subsection{Random pCTMC}

We here outline the details of the pCTMC random generation strategy. Let $r$ denote the number of reactions, i.e. the dimension of the parameter space $\Theta$. We set the maximum allowed number of species to $r+1$. For each species, the upper bound over initial state is set to $10$. For each dimension we generate $5$ models. The parameter space is set to $\Theta = [0.001, 1]^r$ and we sample $5000\cdot r$ parameter values according to a logaritmic hyper-sampling strategy and, for each parameter, we simulate $10$ trajectories. We temporal horizon is set to $T=10$ and $\tau$ of property $\varphi_3$ is set to $20$.

The allowed reactions are of the form: 
\begin{itemize}
\item  $S_i+S_j\to S_k$ sampled with probability $0.25$,
\item  $S_i\to S_j+S_k$ sampled with probability $0.25$,
\item  $S_i\to S_j$ sampled with probability $0.25$,
\item  $S_i\to \emptyset$ sampled with probability $0.125$,
\item  $\emptyset\to S_i$ sampled with probability $0.125$.
\end{itemize}
For each reaction, we randomly sample a type of reaction, according to the associated discrete probability vector, and then we randomly sample the species involved from the the vector $[S_1,\ldots, S_{r+1}]$, assuming a uniform distribution. The number of species actually involved in each model is thus variable. Finally, we check that the overall system is well-defined and non-redundant.  

For each model, we generate $3$ STL properties $\varphi_1$, $\varphi_2$ and $\varphi_3$ by randomly sampling the species to monitor. In order to avoid uninformative properties, we select properties whose satisfaction values over the first $20$ parameter values are non-constant.

\paragraph{Dataset generation.}
Table~\ref{tab:dataset_size_time} shows the chosen dimensions for each generated dataset. In general, the number of observed parameters $N_t$ and the number of observations per point $M_t$ increase proportionally to the dimensionality of $\Theta$.

\begin{table}[ht]
\centering
\adjustbox{max width=\textwidth, max height=\textheight}{
\begin{tabular}{l | c c c c || l | c c c c}
    \toprule
    
    \textbf{Configuration} & \multicolumn{4}{c}{\textbf{Dataset Size}}  & \multicolumn{1}{c}{\textbf{Configuration}} & \multicolumn{4}{c}{\textbf{Dataset Size}}\\
     & $N_t$ & $M_t$ & $N_v$ & $M_v$ & & $N_t$ & $M_t$ & $N_v$ & $M_v$ \\
    \hline
    $(a)$ SIR $\beta$ & $500$ & $50$ & $1k$ & $1k$&$(f)$ PR $k_1$ &  $500$ & $50$ & $1k$ & $1k$\\
    $(b)$ SIR $\gamma$ & $700$ & $50$ & $1k$ & $1k$ &$(g)$ PR $k_p,k_d$& $2500$ & $50$ & $400$ & $1k$ \\
    $(c)$ SIR $\beta,\gamma$ & $2500$ & $50$ & $400$ & $1k$ &$(h)$ PR $k_1,k_2,k_3$ & $8k$ & $20$ & $1k$ & $1k$ \\
    \hline
    $(d)$ PGE $k_2$ & $500$ & $50$ & $400$ & $1k$& $(i)$ PR $k_1,\ldots,k_4$ & $10k$ & $20$ & $4096$ & $1k$\\

    $(e)$ PGE $k_2, k_7$ & $2500$ & $50$ & $400$ & $1k$ &$(l)$ PR $k_p,k_1,\ldots,k_4,k_d$ & $1000k$ & $20$ & $4096$ & $1k$\\
    \hline
    &  & & & &Random pCTMC & $5k\cdot r$ & $10$ & $100\cdot r$ & $1k$  \\
    \bottomrule
\end{tabular}
}
\caption{Size of training ($D_t$) and test ($D_v$). $N_t$ and $N_v$ denote the number of parameter values, while $M_t$ and $M_v$ denote the number of Bernoulli observations for each parameter value. $r$ denotes the dimension of the parameter space. \vspace{-0.5cm}}
\label{tab:dataset_size_time}
\end{table}

\newpage

\section{Statistical Guarantees}\label{app:stat_guar}

\subsection{Inductive Conformal Predictions}\label{app:icp}
In order to provide provable local probabilistic bounds on smMC, we rely on Normalized Inductive Conformal Predictions (NICP)~\cite{Papadopoulos2011ReliablePI,Papadopoulos2014RegressionCP}. 
Let $\err:[0,1]^2\to [0,\infty)$ be the \emph{error function} that measures the absolute error made by a predictor $f$ at a point $(\theta_i,f_\varphi(\theta_i))$. Mathematically, $\err\left(f_\varphi(\theta_i),f(\theta_i)\right) = |f_\varphi(\theta_i)-f(\theta_i)|$, where $f_\varphi$ is the satisfaction function.

We here report and prove Theorem~\ref{thm:nicp} that states how the Bayesian quantification of
uncertainty of smMC can be exploited and used as the normalizing function of a NICP
approach, that in turn will provide us with point-specific statistical guarantees over
the error coverage.

\begin{theorem-non}\label{thm:nicp_app}
Consider a pCTMC $\mathcal{M}_\theta$, an STL formula $\varphi$, a dataset $D_t$ used to infer the smMC posterior distribution $q(f)$ over functions $f$ that approximates the satisfaction function $f_\varphi$, and a calibration set $D_c$. Let the normalizing function $u(\theta)$ be the standard deviation $\sigma_{q(f)}(\theta)$ of the posterior $q(f)$. Then, for a significance level $\varepsilon\in (0,1)$

\begin{equation}\label{eq:thm_app}
    Pr_{(\theta,L)\sim p(\theta,f_\varphi(\theta))}\Big(
 \delta\big(f_\varphi(\theta),\ \mathbb{E}_{f\sim q(f)}[f(\theta)]\big)\le\tilde{\tau}_\varepsilon\cdot \sigma_{q(f)}(\theta)
    \Big) \ge 1-\varepsilon,
\end{equation}
where $\tilde{\tau}_\varepsilon$ is the $(1-\varepsilon)$-quantile of the scores $\tilde{\tau}_c$ computed over the calibration set $D_c$:
\begin{equation}
\tilde{\tau}_c = \Bigg\{\frac{\delta\big(f_\varphi(\theta_i),f(\theta_i)\big)}{u(\theta_i)}\ \Big| \ (\theta_i,f_\varphi(\theta_i))\in D_c\Bigg\},
\end{equation}
\end{theorem-non}

\begin{proof}
Inductive Conformal Predictions (ICP)~\cite{vovk2005algorithmic} assign
reliable confidence measures to predictions without assuming anything more than that the data are independent and identically distributed (i.i.d.). ICP are built on top of traditional machine learning algorithms and accompany each of their predictions with valid measures of
confidence. Unlike Bayesian methods, ICPs do not require any further assumptions about
the data distribution.  Even if the traditional algorithm to which a CP is applied to makes some extra assumptions that are not true for a particular data set, the validity of the predictive regions produced by the ICP will not be affected. The resulting predictive regions might be
uninteresting, but they will still be valid, as opposed to the misleading regions produced
by Bayesian methods. In order to apply ICP to a traditional algorithm one has to develop
a nonconformity measure based on that algorithm. This measure evaluates the difference
of a new example from a set of old examples. Many different nonconformity measures can be constructed for each traditional algorithm and each of those measures defines a different CP. This difference does not affect the validity of the results produced by the CPs, it only affects their efficiency, i.e. the width of the prediction regions.
For detailed proof, we remind the interested reader to~\cite{vovk2005algorithmic,Papadopoulos2014RegressionCP,Papadopoulos2011ReliablePI}.
\end{proof}

\paragraph{Chernoff bounds for SMC.} Theorem~\ref{thm:nicp} provides guarantees over the error made w.r.t. to the ground truth target, the satisfaction function $f_\varphi$. Unfortunately, this quantity cannot be observed exactly. To practically compute a bound, we resort to SMC and estimate it as
$\bar{L}_i: = \tfrac{1}{M_t}\sum_{j_=1}^{M_t}\ell_i^j$, evaluating the nonconformity scores $\tau_c$ as a distance from $\bar{L}_i$ rather than $f_\varphi(\theta_i)$. This introduces an additional error that can be taken into account by combining~\eqref{eq:thm_main} with standard Chernoff bounds for  SMC. Mathematically, let $\bar{f}(\theta):= \mathbb{E}_{f\sim q(f)}[f(\theta)]$, we know that 
$$\err \big(f_\varphi(\theta_i), \bar{f}(\theta_i)\big) \le \err \big(\bar{L}_i,\bar{f}(\theta_i)\big) + \err \big(f_\varphi(\theta_i), \bar{L}_i\big)$$ and that $Pr\Big(\err \big(\bar{L}_i,\bar{f}(\theta_i)\big)\le T_1(\theta_i)\Big)\ge 1- \varepsilon_1$, where $T_1(\theta_i) := \tilde{\tau}_{\varepsilon_1}\cdot \sigma_{q(f)}(\theta_i)$. The Chernoff bound for SMC~\cite{Legay2010StatisticalMC} allows the user to specify a point specific error bound $T_2(\theta)$ and a significance level $\varepsilon_2\in (0,1)$ such that $Pr\Big(\err \big(f_\varphi(\theta_i), \bar{L}_i\big)\le T_2(\theta_i)\Big)\ge 1-\varepsilon_2$. The significance level $\varepsilon_2\in (0,1)$ is related to the number of simulations $M_i$ needed to reach a desired error bound $T_2(\theta_i)$ by the relation $\varepsilon_2 = 2\cdot \exp(-2M_iT_2^2)$.  
This information can be combined with a union bound, so that
$$
Pr\Big(\err \big(f_\varphi(\theta_i), f(\theta_i)\big) \le \tau_{\varepsilon_1}+\tau_{\varepsilon_2}\Big)\ge 1-(\varepsilon_1+\varepsilon_2). 
$$
In practice, for a parameter $\theta_i$ with $500$ observations, as the maximum number used in our experiments both for $D_t$ and for $D_c$, the error bound $T_2(\theta_i)$ for $\varepsilon_2 = 0.05$ is around $0.06$ (see Fig.~\ref{fig:unc_comp_2}).


\subsection{PAC-Bayes bound}\label{app:pac}
We follow the easy-to-follow notation introduced in~\cite{Alquier2021UserfriendlyIT}. Let the \emph{generalization error} $\mbox{gen}(f)$ and the \emph{empirical error} $\mbox{emp} (f)$ of a predictor $f$ be defined as
\begin{equation*}
\displaystyle
    \mbox{gen}(f)= \mathbb{E}_{(\theta,L)\sim p(\theta,L)}\left[\err\left(\bar{L},f(\theta)\right)\right]; \qquad
    \mbox{emp} (f) = \frac{1}{N_t}\sum_{i=1}^{N_t} \err(\bar{L}_i,f(\theta_i)),  
\end{equation*}
where $p(\theta,L)$ is the unknown data-generating distribution. We aim at computing an empirical quantity that upper bounds $\mbox{gen}(f)$ with a given confidence.
Let $\mathcal{P}(F)$ be the set of all probability distributions over space $F$ of functions of the form $\Theta\to [0,1]$ and fix a ``prior'' distribution $\pi\in\mathcal{P}(F)$ ($\pi$ should not depend on $D_t$). The 
Catoni's PAC-Bayes bound~\cite{Catoni2004APA} states that for any $\lambda > 0$ and for any $\epsilon \in (0,1)$,
\begin{equation}\label{eq:pac_bound_app}
    Pr\Bigg(\forall \rho\in\mathcal{P}(F),\ \mathbb{E}_{f\sim \rho}[\mbox{gen}(f)]\le \mathbb{E}_{f\sim\rho}[\mbox{emp}(f)]+\frac{\lambda C^2}{8N_t}+\frac{KL[\rho||\pi]+\log\tfrac{1}{\epsilon}}{\lambda}\Bigg)\ge 1-\epsilon,
\end{equation}
where $C$ is the upper bound of the bounded error function.
As~\eqref{eq:pac_bound_app} holds for any $\rho\in\mathcal{P}(F)$, it holds for the smMC distribution $q(f)$ that approximates the posterior distribution $p(f|D_t)$. Moreover, the second part of the bound can be made as tight as possible by minimizing it w.r.t. $\lambda$.
The resulting PAC-Bayes bound for the expected generalization error can be expressed as 

\begin{equation}\label{eq:pac_bound_min}
    Pr\Bigg(\mathbb{E}_{f\sim q(f)}[\mbox{gen}(f)]\le \mathbb{E}_{f\sim q(f)} [\mbox{emp}(f)]+\frac{\lambda_* }{8N_t}+\frac{KL[q||\pi]+\log\tfrac{1}{\epsilon}}{\lambda_*}\Bigg)\ge 1-\epsilon,
\end{equation}
where $C=1$ and 
$\lambda_* = \sqrt{\frac{8n}{C^2}\left(KL[q||\pi]+\log\tfrac{1}{\epsilon}\right)}$.



The bound should be adjusted in order to be applicable to all models: EP-GP, SVI-GP and SVI-BNN.
In particular, the main difference are the following:
\begin{itemize}
    \item \textbf{GP}: inference is performed over real-valued functions $g:\Theta\to\mathbb{R}$. The distribution $\pi(g(\cdot))$ is defined as a zero-mean GP with kernel $\tilde{k}$ and $q(g(\cdot))$ is the approximation of the intractable posterior and is, in turn, a GP. The KL divergence between these two GP is equivalent to the KL divergence between the GP evaluated over the training points. Mathematically, $$KL\left[q(g(\cdot))||\pi(g(\cdot))\right] = KL\left[q(g_t)||\pi(g_t)\right],$$ where both $q(g_t)$ and $\pi (g_t) = \mathcal{N}(0_{N_t}, \tilde{K}_{N_tN_t})$ are Gaussians over $\mathbb{R}^{N_t}$. This result is proved in~\cite{Seeger2003PACBayesianGE}.
    \item \textbf{BNN}: inference is performed or the weights $\mathbf{w}$ of the neural network $f_\mathbf{w}:\Theta\to [0,1]$. $\pi(\mathbf{w})$ is a standard Gaussian over $\mathbb{R}^{|\mathcal{W}|}$, where $\mathcal{W}$ denotes the weights' space, $q_\psi (\mathbf{w})$ is also Gaussian over $\mathbb{R}^{|\mathcal{W}|}$. Weights are independently distributed, thus $KL[q(\mathbf{w})||\pi(\mathbf{w})] = \sum_{i=1}^{|\mathcal{W}|}KL[q(w_i)||\pi(w_i)]$.
\end{itemize}
The KL divergence between Gaussian distributions has an analytic formulation and thus it can be easily computed.

\begin{figure}[!t]
    \centering
    \includegraphics[width=\textwidth]{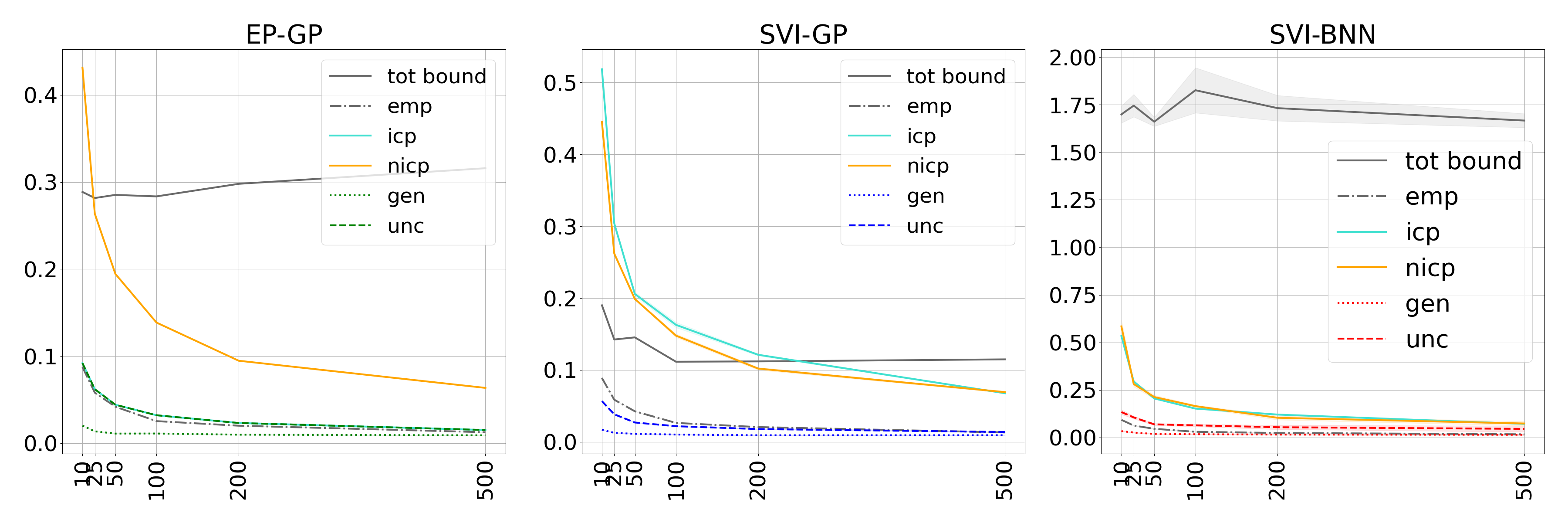}
    \caption{Configuration $(a)$: comparison of PAC-Bayes bound (grey), ICP bound (cyan) and NICP bound (orange) for EP-GP (left), SVI-GP (middle) and SVI-BNN (right). All bounds are over the expectation over the entire parameter space. The number of observed parameters is $N_t = 500$, the number of observations per point, $M_t$, increases over the horizontal axis. Plots show the average and the $95\%$ confidence interval over 5 runs of training. The dotted, dash-dot and dashed lines denote the generalization error, the empirical error and the average uncertainty area (Unc). The expected generalization error (gen) is estimated over $D_v$. \vspace{-0.5cm}}
    \label{fig:pac_500}
\end{figure}

\newpage

\section{Comparison of the uncertainty estimates}\label{app:unc_estim}

\begin{figure}[!b]
    \centering
    \includegraphics[width=\textwidth]{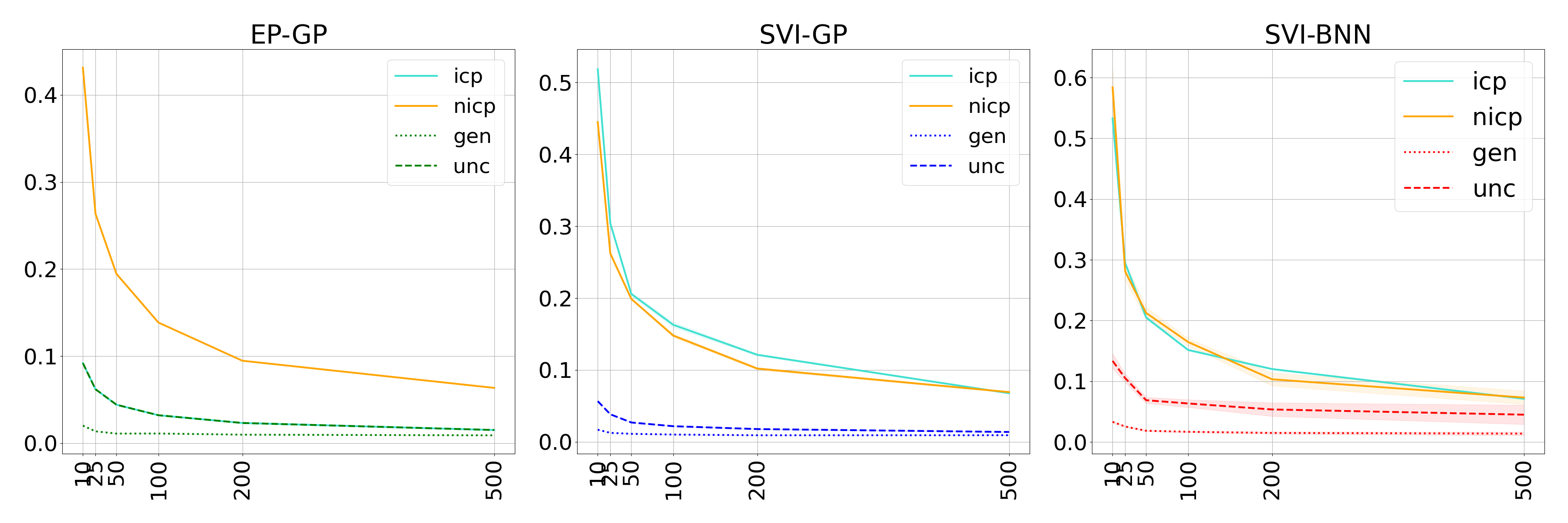}
    \caption{Configuration $(a)$: comparison of the Bayeasian uncertainty quantification with ICP (cyan) and normalized ICP (orange) bound for EP-GP (left), SVI-GP (middle) and SVI-BNN (right). The number of observed parameters is $N_t = 500$, the number of observations per point, $M_t$, increases over the horizontal axis. Plots show the average and the $95\%$ confidence interval over 5 runs of training. The dotted, dash-dot and dashed lines denote the expected generalization error and the average uncertainty area (Unc). The expected generalization error (gen) is estimated over $D_v$. \vspace{-0.5cm}}
    \label{fig:avg_unc_500}
\end{figure}

\begin{figure}
    \centering
    \includegraphics[width=\textwidth]{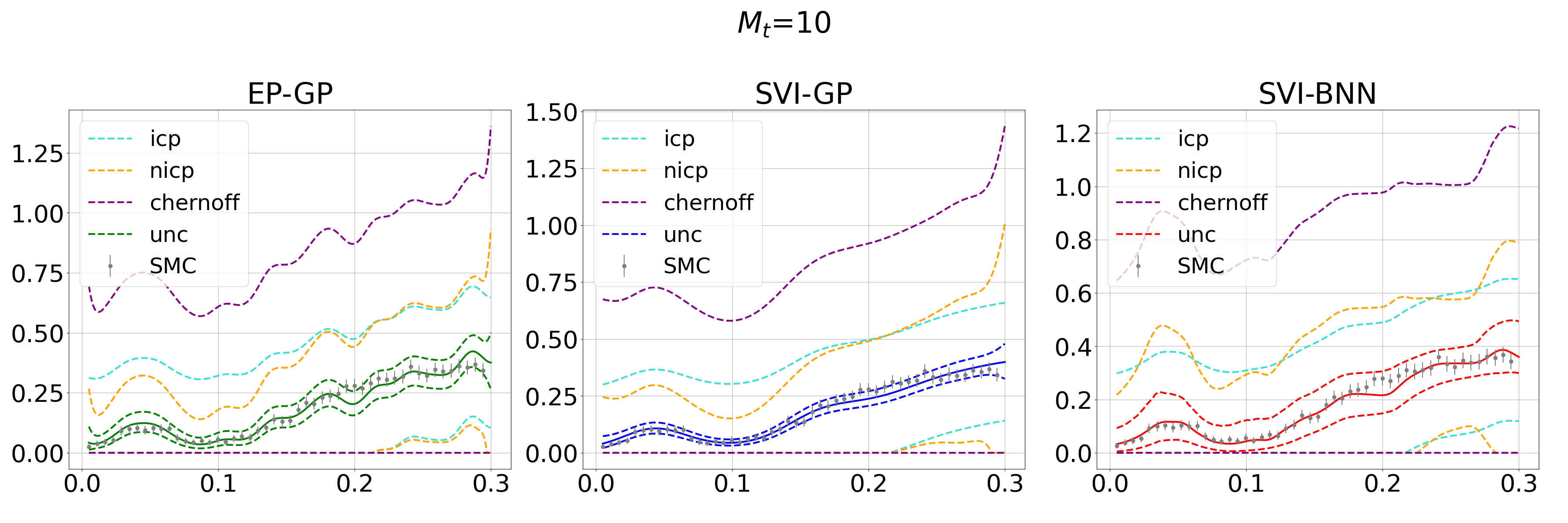}
    \includegraphics[width=\textwidth]{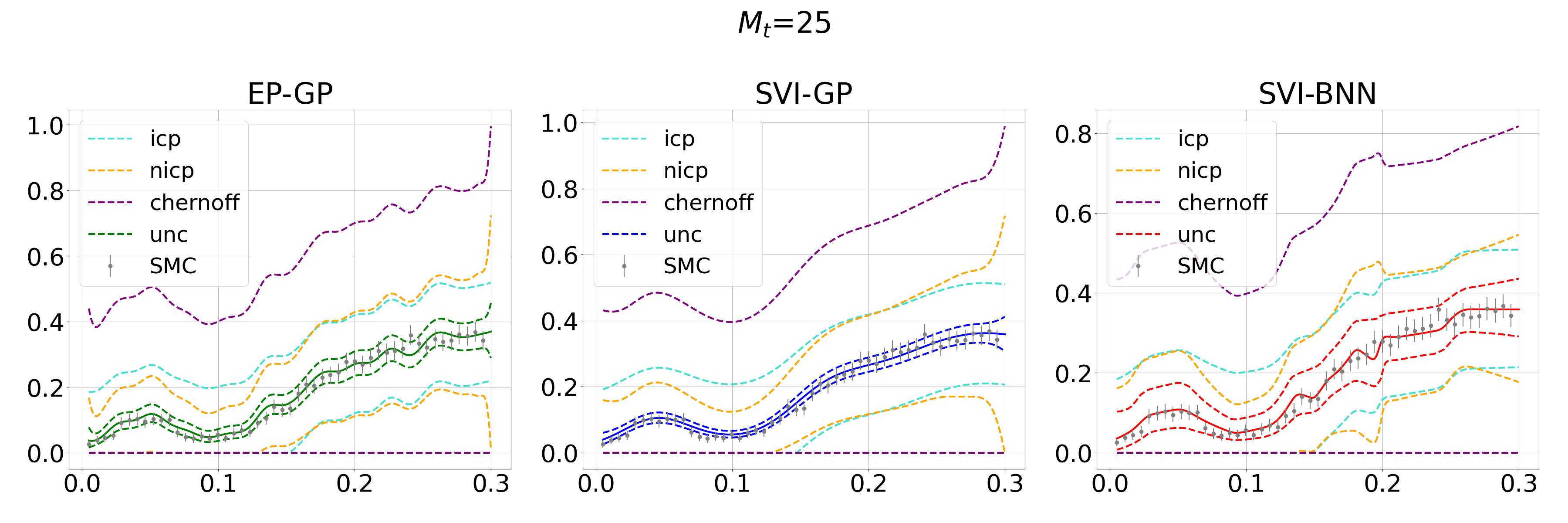}
    \includegraphics[width=\textwidth]{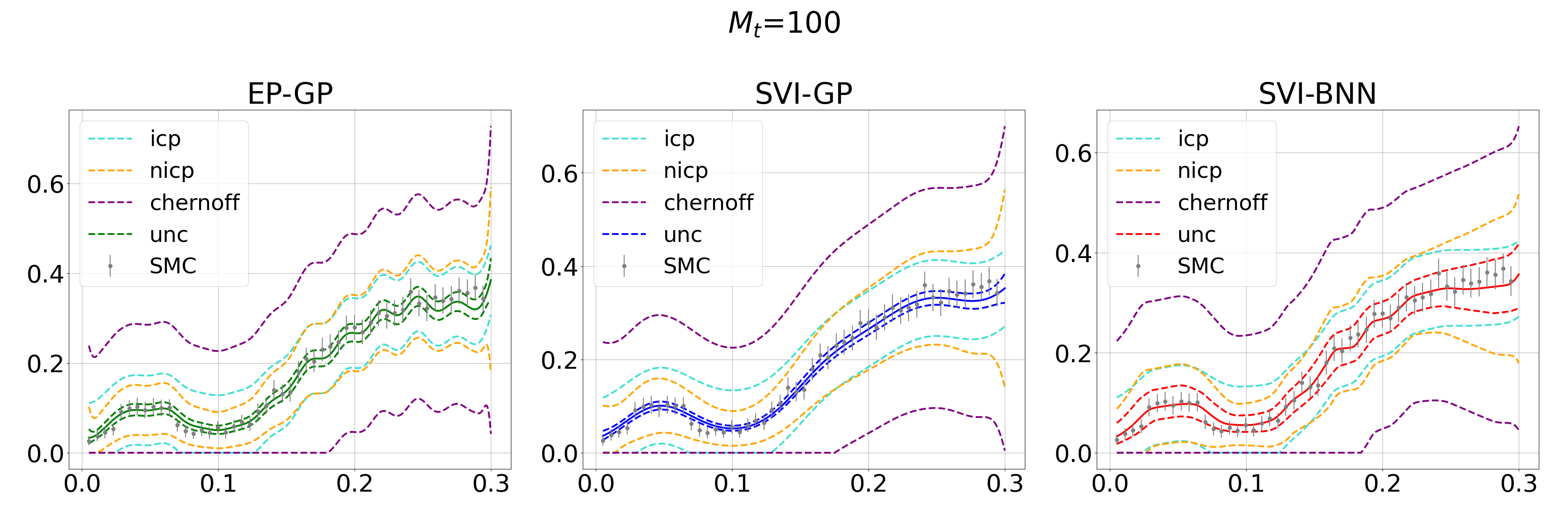}
    \includegraphics[width=\textwidth]{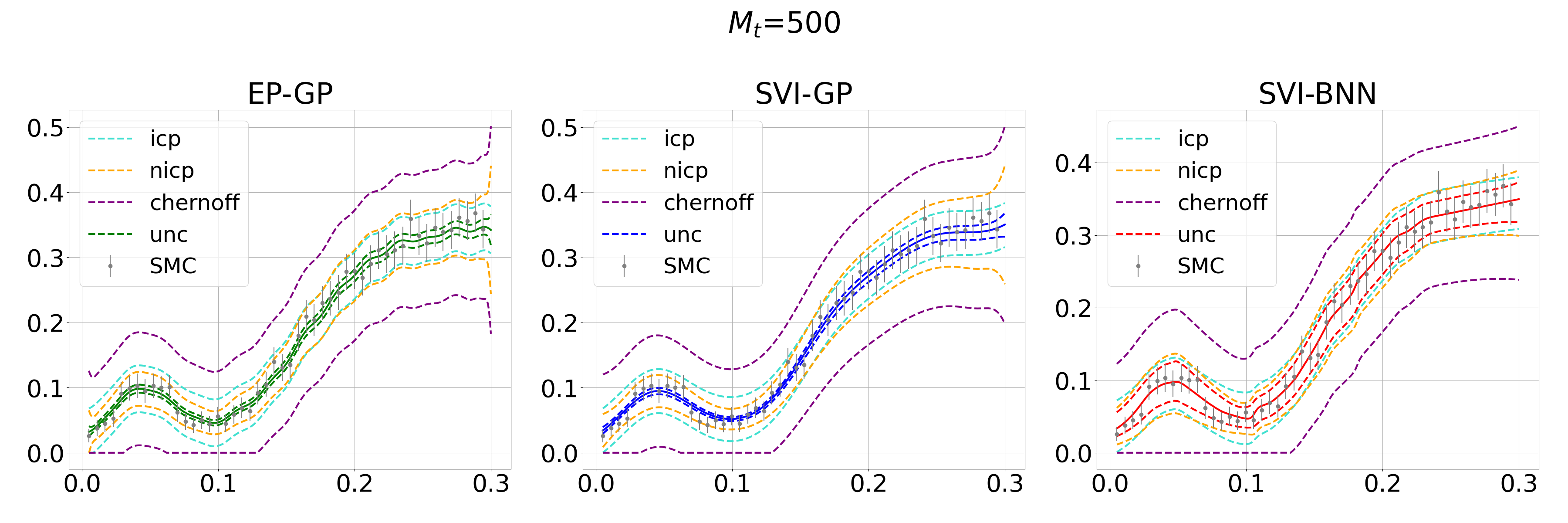}

    \caption{Comparison of the point-wise Bayesian estimate of uncertainty with those of ICP (cyan) and of the NICP (orange) for increasing values of $M_t$. Chernoff bound correction (purple) provide guaranteed $90\%$ coverage over the exact satisfaction function.}
    \label{fig:unc_comp_2}
\end{figure}

Comparison of the average widths over $D_v$ as a function of $M_t$ (Fig.~\ref{fig:avg_unc_500}) and comparison of the point-wise bounds as a function of $\Theta$ (Fig.~\ref{fig:unc_comp_2}). In particular, the Bayesian estimate of uncertainty is compared with those of ICP (cyan) and of the normalized ICP (orange) for increasing values of $M_t$. Grey error bars denote the SMC uncertainty.

\section{Additional Plots}\label{app:plots}

\begin{figure*}[!b]
    \centering

    \includegraphics[width=\linewidth]{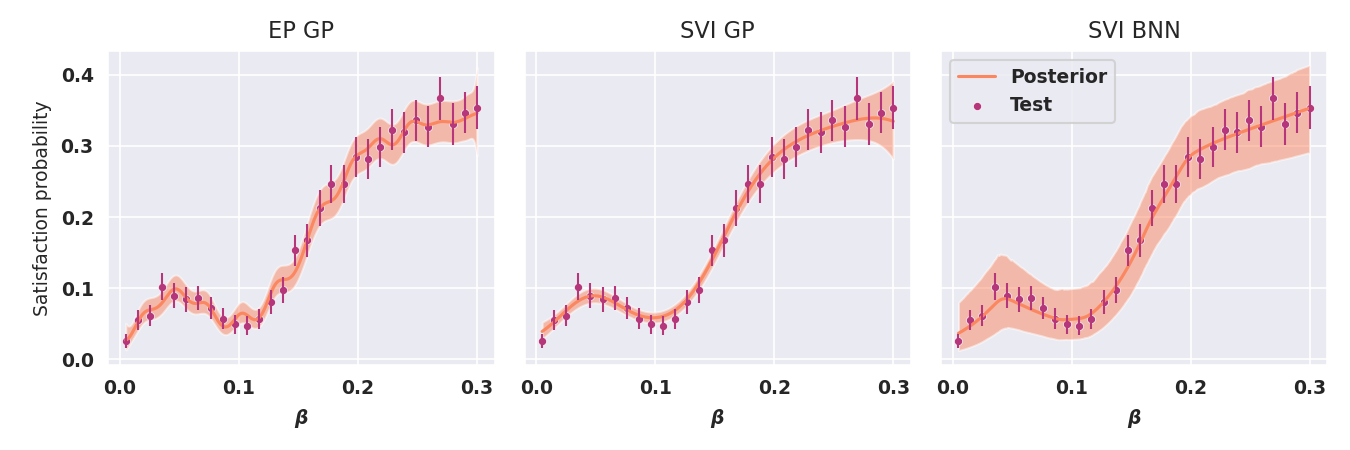}
    \caption{Configuration $(a)$}
    \label{fig:SIR_beta}
    \includegraphics[width=\linewidth]{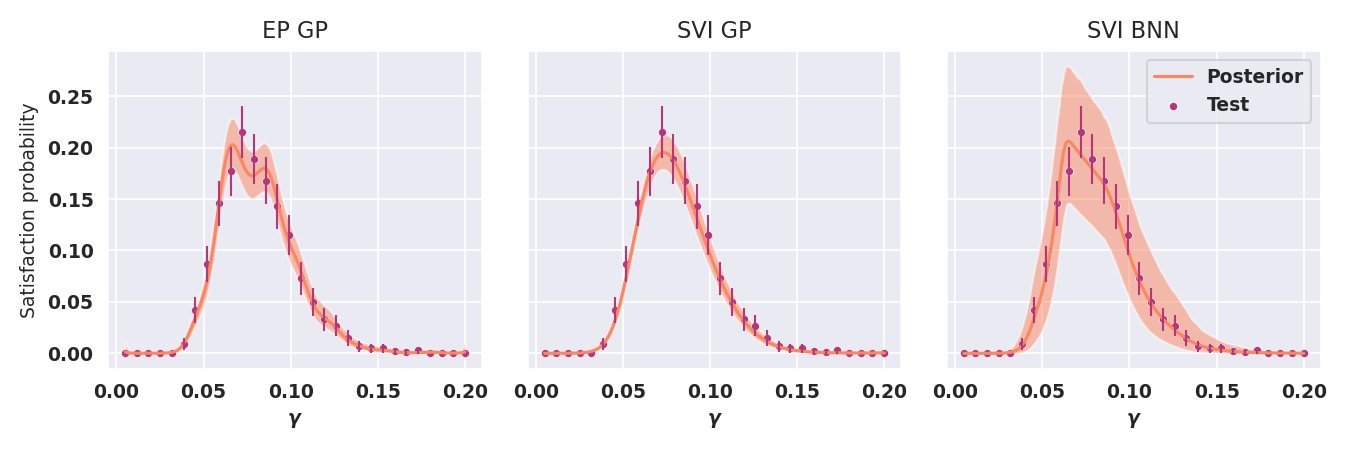}
    \caption{Configuration $(b)$ 
    }\label{fig:SIR_gamma}

    \includegraphics[width=\linewidth]{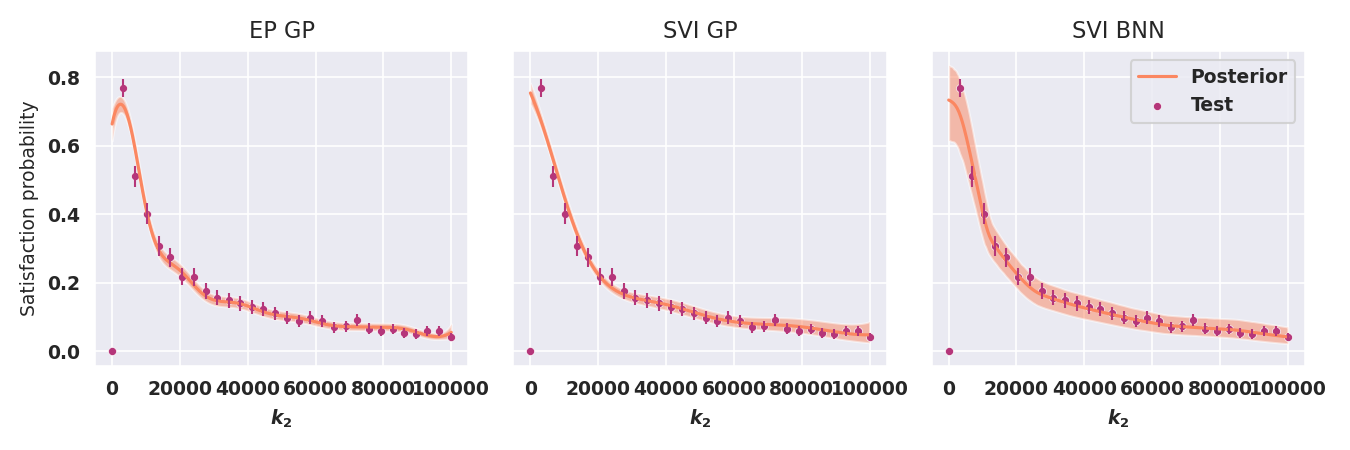}
    \caption{Config. $(d)$: satisfaction probability estimated by EP-GPs, SVI-GPs and SVI-BNNs ($1k$ posterior samples) and true satisfaction probability on $30$ equispaced points from the test set $D_v$, with $95\%$ confidence intervals around the mean. 
    }    \label{fig:PrGeEx_k2}
\end{figure*}

\begin{figure*}[!t]
    \centering
        \includegraphics[width=\linewidth]{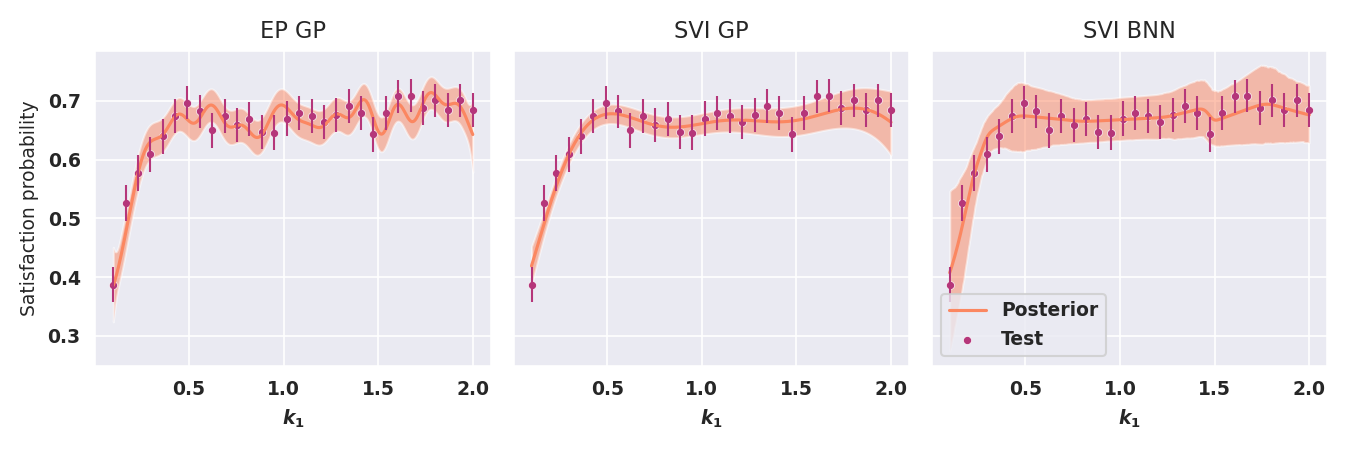}
    \caption{Configuration $(f)$: satisfaction probability estimated by EP-GPs, SVI-GPs and SVI-BNNs ($1k$ posterior samples) and true satisfaction probability on $30$ equispaced points from the test set $D_v$, with $95\%$ confidence intervals around the mean. }\label{fig:PhosRelay_k1}
        \includegraphics[width=\linewidth]{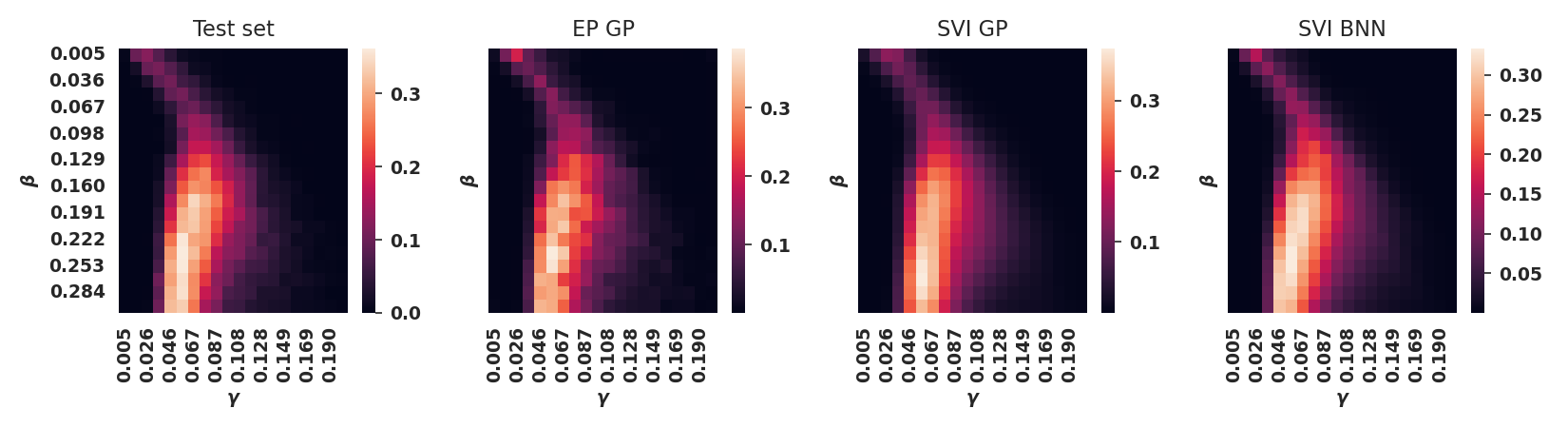}
    \caption{True satisfaction probability is compared to the satisfaction probability estimated by EP-GPs, SVI-GPs and SVI-BNNs on the test set for configuration (c). 
    SVI-BNNs are evaluated on $1k$ posterior samples.\vspace{-0.5cm}}
    \label{fig:SIR_betagamma}
    
    \includegraphics[width=\linewidth]{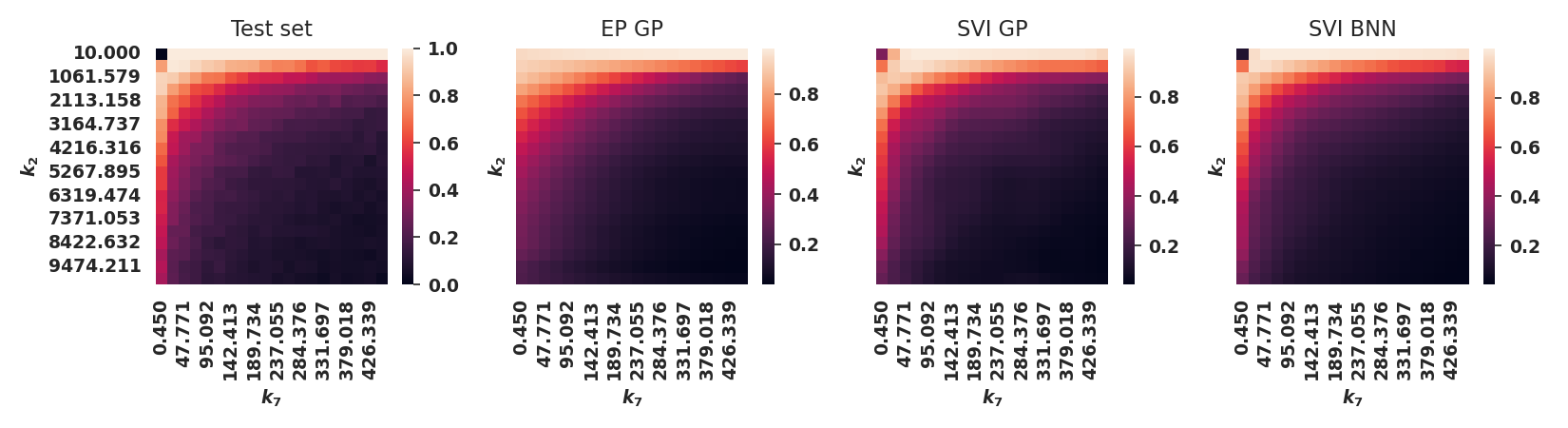}
    \caption{True satisfaction probability is compared to the satisfaction probability estimated by EP-GPs, SVI-GPs and SVI-BNNs ($1k$ posterior samples) on the test set for configuration (e).
    }
    \label{fig:PrGeEx_k2k7}

    \includegraphics[width=\linewidth]{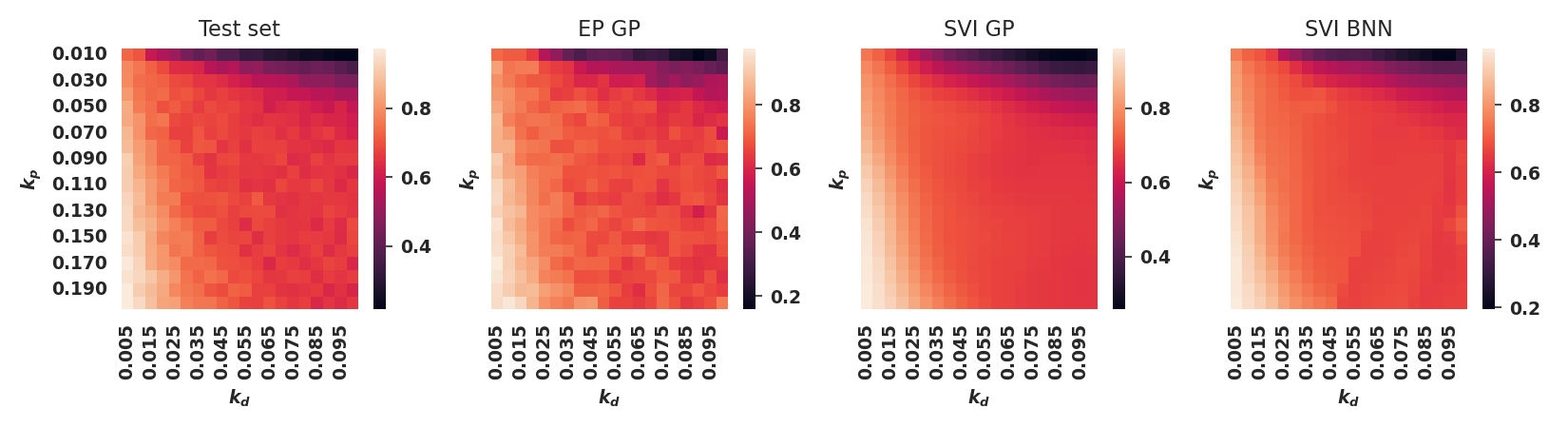}
    \caption{True satisfaction probability is compared to the satisfaction probability estimated by EP-GPs, SVI-GPs and SVI-BNNs ($1k$ posterior samples) on the test set for configuration (g). 
    }
    \label{fig:PhosRelay_kpkd}
\end{figure*}

Qualitative analysis of the performance of SV-smMC in reconstructing the satisfaction function from SMC observations. For visualization purposes, only one-dimensional (Fig.~\ref{fig:SIR_beta}-\ref{fig:PhosRelay_k1} for $(a)$, $(b)$, $(d)$ and $(f)$ respectively) and two-dimensional (Fig.~\ref{fig:SIR_betagamma}-\ref{fig:PhosRelay_kpkd} for $(c)$, $(e)$ and $(g)$) biological case studies are shown.  

\end{document}